\documentclass[11pt,a4paper]{article}
\usepackage[hyperref]{emnlp2020}
\usepackage{times}
\usepackage{latexsym}

\usepackage{microtype}

\usepackage{amsfonts,amsmath,amssymb,amsthm}
\usepackage{bm,nicefrac}
\usepackage{bbding}
\usepackage{multirow}
\usepackage{subfigure}
\usepackage{graphicx}
\usepackage{comment}

\aclfinalcopy % Uncomment this line for the final submission
\title{Graph-to-Tree Neural Networks \\ for Learning Structured Input-Output Translation \\ with Applications to Semantic Parsing and Math Word Problem}

\author{
  Shucheng Li$^*$$^{\dag}$, 
  Lingfei Wu\thanks{\ \, authors contributed equally to this research.}\ \,$^{\ddag}$, 
  Shiwei Feng$^{\dag}$, 
  Fengyuan Xu$^{\dag}$, 
  Fangli Xu$^{\S}$, 
  Sheng Zhong$^{\dag}$ \\
$\dag$National Key Lab for Novel Software Technology, Nanjing University, China\\ 
$\ddag$IBM Thomas J. Watson Research Center, Yorktown Heights, NY, USA\\
$\S$Squirrel AI Learning, New Jersey, USA\\
\small {\tt $\dag$\{shuchengli,fengyuan.xu,swfeng98,sheng.zhong\}@smail.nju.edu.cn}\\
\small {\tt $\ddag$lwu@email.wm.edu}, \
\small {\tt $\S$lili@yixue.us}
}

\date{}

\begin{document}
\maketitle
\begin{abstract}
The celebrated Seq2Seq technique and its numerous variants achieve excellent performance on many tasks such as neural machine translation, semantic parsing, and math word problem solving. However, these models either only consider input objects as sequences while ignoring the important structural information for encoding, or they simply treat output objects as sequence outputs instead of structural objects for decoding. In this paper, we present a novel Graph-to-Tree Neural Networks, namely Graph2Tree consisting of a graph encoder and a hierarchical tree decoder, that encodes an augmented graph-structured input and decodes a tree-structured output. In particular, we investigated our model for solving two problems, neural semantic parsing and math word problem. Our extensive experiments demonstrate that our Graph2Tree model outperforms or matches the performance of other state-of-the-art models on these tasks.
\end{abstract}

\section{Introduction}
\label{sec:intro}

Learning general functional dependency between arbitrary input and output spaces is one of the key challenges in machine learning.
While many efforts in machine learning have mainly
focused on designing flexible and powerful input representations for solving classification or regression problems,
many applications require researchers to design novel models that can deal with complex structured inputs and outputs,
such as graphs, trees, sequences, or sets. In this paper,
we consider the general problem of learning a mapping between a graph input $G \in \mathcal{G}$ and a tree output $T \in \mathcal{T}$,
based on a training sample of structured input-output
pairs $(G_1,T_1),...,(G_n,T_n) \in \mathcal{G} \times \mathcal{T}$ drawn from some fixed but unknown probability distribution.

Such learning problems often arise in a variety of applications,
ranging from semantic parsing, to math word problem, label sequence learning,
and supervised grammar learning, to name just a few. As shown in Fig. \ref{tab:sp_mwp_sample}, finding the parse tree of a sentence involves a structural dependency among the labels in the parse tree;
generating a mathematical expression of a math word problem involves a hierarchical dependency between math logical operations and the numbers.
Conventionally, there have been efforts in generalizing kernel methods to predict structured and inter-dependent variables in a
supervised learning setting \cite{tsochantaridis2005large,altun2004gaussian,joachims2009predicting}.
%like SVMs \cite{tsochantaridis2005large} and logistic regression \cite{altun2004gaussian}.
% ,altun2006maximum,joachims2009predicting

% table 1

\begin{table}[t]
% \footnotesize
\scriptsize
\centering
% \caption{Examples of structured input and output of semantic parsing (SP) and math word problem (MWP). For inputs, we consider parsing tree augmented sequences to get structural information. For outputs, they are naturally a hierarchical structure with some structural meaning symbols like brackets.}
\newcommand{\Bd}[1]{\textbf{#1}}
% \vspace{-2mm}
\begin{tabular}{|c|p{6.0cm}|}
\hline
\multirow{6}*{SP} & \textbf{Text Input:} \\
~ & what jobs are there for web developer who know 'c++' ? \\ [0.3cm]

~ & \textbf{Structured output:} \\
~ & answer( A , ( job ( A ) , title ( A , W ) , const ( W , 'Web Developer' ) , language ( A , C ) , const ( C , 'c++' ) ) ) \\
\hline

\multirow{7}*{MWP} & \textbf{Text input:}\\
% ~ & 0.5 of the cows are grazing grass .  0.75 of the remaining are sleeping and 9 cows are drinking water from the pond .  find the total number of cows .\\[0.4cm]
~ & 0.5 of the cows are grazing grass .  0.25 of the cows are sleeping and 9 cows are drinking water from the pond .  find the total number of cows .\\[0.3cm]

~ & \textbf{Structured output:}\\
~ & ( ( 0.5 * x ) + ( 0.25 * x ) ) + 9.0 = x\\
\hline
\end{tabular}
\caption{Examples of structured input and output of semantic parsing (SP) and math word problem (MWP). For inputs, we consider parsing tree augmented sequences to get structural information. For outputs, they are naturally a hierarchical structure with some structural meaning symbols like brackets.}
\label{tab:sp_mwp_sample}
% \vspace{-4mm}
\end{table}

Recently, the celebrated Sequence-to-Sequence technique (Seq2Seq) and its numerous variants \cite{sutskever2014sequence,bahdanau+al-2014-nmt,luong-etal-2015-effective} achieve excellent performance in neural machine translation. Encouraged by the success of Seq2Seq model, there is a surge of interests in applying Seq2Seq models to cope with other tasks such as developing neural semantic parser \cite{dong-lapata-2016-language} or solving math word problem \cite{ling-etal-2017-program}. However, the two significant challenges making a Seq2Seq model ineffective in these tasks are that, i) for the natural text description input, it often entails some hidden syntactic structure information such as dependency, constituency tree or even semantic structure information like AMR parsing tree; ii) for the meaningful representation output, it typically contains abundant information in a structured object like a parsing tree or a mathematical equation.

Inspired by these observations, in this work, we propose a Graph-to-Tree neural networks, namely Graph2Tree consisting
of a graph encoder and a hierarchical tree decoder, which leverages the structural information of both source graphs and
target trees. In particular, our Graph2Tree model learns the mapping from a structured object such as a graph to another
structured object such as a tree. In addition, we also observe that the structured object translation typically follows
a modular procedure, which translates the individual sub-graph in the source graph into the corresponding target one in target tree output, and then compose them to form the final target tree.

Therefore, we design a workflow to align with this procedure: our graph encoder first learns from an input graph that is constructed from the various inputs such as combining both a word sequence and the corresponding dependency or constituency tree, and then our tree decoder generates the tree object from the learned graph vector representations to explicitly capture the compositional structure of a tree. In particular, we present a novel Graph2tree model with a separated attention mechanism to jointly learn a final hidden vector of the corresponding graph nodes in order to align the generation process between a heterogeneous graph input and a hierarchical tree output.

To demonstrate the effectiveness of our model, we perform experiments on two important tasks -- Semantic Parsing and Math Word Problem. First, we compare our approach against several neural network approaches on the Semantic Parsing task. Our experimental results show that our Graph2Tree model could outperform or match the performance of other state-of-the-art models on three standard benchmark datasets. Second, we further compare our approach with existing recently developed neural approaches on the math word problem and our results clearly show that our Graph2Tree model can achieve state-of-the-art performance compared to other baselines that use many task-specific techniques. 
We believe our Graph2Tree model is a solid attempt for learning structured input-output translation.
%and we consider developing general Graph2Graph model as next future work.

\section{Related Works}

\subsection{Graph Neural Networks}
The graph representation learning recently attracted a lot of attention and interest from both academia and industry.
One of the most important research lines is the semantic embedding learning of graph nodes or edges based upon the power of
graph neural networks (GNNs)~\cite{li2016gated,kipf2017semi,Velickovic2017GraphAN,gilmer2017neural,hamilton2017inductive}.

% On one hand, various Sequence-to-Graph or Graph-to-Sequence algorithms have been proposed to handle the structured
% inputs, structured outputs or both of them. For instance,  Peng et al.~\cite{peng2018amr} proposed a AMR graph generation from the text
% sequence inputs, while some other researchers \cite{xu2018graph2seq,song2018graph} focused on translating graphs into
% sequences. On the other hand, some research works proposed the Tree-to-Tree \cite{chen2018tree} or Graph-to-Graph
% \cite{guo2018deep} neural networks for targeted application scenarios.
Encouraged by the recent success in GNNs, various Sequence-to-Graph ~\cite{peng2018amr} or Graph-to-Sequence models \cite{xu2018graph2seq,xu2018exploiting, xu2018sql, beck-etal-2018-graph,chen2019reinforcement} have been proposed to handle the structured inputs, structured outputs or both of them, i.e. generating AMR graph generation from the text
sequence. More recently, some researchers proposed the Tree-to-Tree \cite{chen2018tree}, Graph-to-Tree
\cite{YinNABG19} and Graph-to-Graph \cite{guo2018deep} neural networks for targeted application scenarios.

However, these works are designed exclusively for specific downstream tasks like program translation or code edit. Compared to them, our proposed Graph2Tree neural network with novel design of graph encoder and tree decoder does not rely on any
specific downstream task assumption. Additionally, our Graph2Tree is the first generic neural network translating graph
inputs into tree outputs, which may have numerous applications in practice.

\subsection{Neural Semantic Parsing}
Semantic parsing is the task of translating natural language utterances into machine-interpretable meaning representations like
logical forms or SQL queries. Recent years have witnessed a surge of interests in developing neural semantic
parsers with sequence-to-sequence models. These parsers have achieved promising results
%~\cite{jia-liang-2016-data}.
~\cite{jia-liang-2016-data,dong-lapata-2016-language,ling-etal-2016-latent}.
Due to the fact that the meaning representations are usually structured objects (e.g. tree structures), many efforts
have been devoted to develop structure-oriented decoders, including
tree decoders
% \cite{dong-lapata-2016-language},
\cite{dong-lapata-2016-language,alvarez2016tree},
grammar constrained decoders
% \cite{yin-neubig-2017-syntactic},
\cite{yin-neubig-2017-syntactic,yin-etal-2018-structvae,jie-lu-2018-dependency,dong-lapata-2018-coarse},
action sequences for semantic graph generation \cite{chen-etal-2018-sequence},
and modular decoders based on abstract syntax trees
\cite{rabinovich-etal-2017-abstract}.
% \cite{rabinovich2017abstract,yu-etal-2018-syntaxsqlnet}.
However, those approaches could potentially be further improved because they only consider the word sequence information
and ignore other rich syntactic information, such as dependency or constituency tree, available at the encoder side.

Researchers recently attempted to leverage of the power of GNNs in various NLP tasks, including the
neural machine translation
\cite{bastings-etal-2017-graph,beck-etal-2018-graph},
conversational machine reading comprehension \cite{chen2019graphflow},
and AMR-to-text \cite{song-etal-2018-graph}.
Specifically in the semantic parsing field, a general Graph2Seq model \cite{xu2018exploiting} is proposed to incorporate
these dependency and constituency trees with the word sequence and then create a syntactic graph as the encoding input.
However, this approach simply treats a logical form as a sequence, neglecting the abundant information in a structured
object like tree in the decoder architecture. Therefore, we present the Graph2Tree model to utilize the structure
information in both structured inputs and outputs.

%GNNs started to be used in semantic parsing, which showed promising results compared non-GNNs based meethods. However, in order to apply GNNs, the first step is to construct a graph that incorporates different structural information such as word sequence, dependent parsing tree, and constituency tree. However, there are many ways to construct a graph and it has no or less studies to investigate the impact of the different graph constructions to the performance of final tasks.

\subsection{Math Word Problems}
% when writing RW:
% write the research line and connections between prior works
% what's limitations in previous works and how do you solve it.

The math word problem is the task of translating the short paragraph (typically consisting with multiple short sentences) into succinct mathematical equations.
To solve a math word problem illustrated in Table \ref{tab:sp_mwp_sample}, 
%algorithms should precisely model implicit and explicit quantities and relationships in the input sentences.
% Traditional approaches focus on statistical machine-learning methods to generate numeric answer expression by by mapping verbs in problems text to categories
traditional approaches focus on generating numeric answer expressions by mapping verbs in problems text to categories
\cite{hosseini-etal-2014-learning} or by generating templates from problem texts \cite{kushman-etal-2014-learning}. However, these approaches either need additional hand-crafted annotations for problem texts or are limited to a set of predefined equation templates.

% solve simple math word problems (only including addition and subtraction) by mapping verbs in problems text to categories and then update impact on the world states to generate numeric answer expression.
% \cite{kushman2014learning} handle math word problems by generating templates from problem texts and then instantiating them with numbers and nouns in source sentences.

% Considering the natural hierarchical structure of mathematical expressions, some semantic parsing-based methods \cite{koncel2015parsing} \cite{roy2016solving} \cite{roy2017unit} are proposed to generate the equation representation in a bottom-up manner.
% Firstly, \cite{roy2016solving} define an expression tree to decompose the problem mapping from texts to arithmetic equations to some prediction problems like predicting proper operation types (addition, subtraction, multiplication and division) between quantities from the problem texts. Following this work, \cite{koncel2015parsing} applies linear programming to seek for all possible equation trees and \cite{roy2017unit} proposed Unit Dependency Graph (UDG) to model the relationship between number units in the problem texts, which need to be predicted in a joint inference framework.

Inspired by the great success of Seq2Seq models in Neural Machine Translation, deep-learning based methods are
intensively explored by researchers in the equation generation~\cite{wang-etal-2017-deep,ling-etal-2017-program,li-etal-2018-multi-head,li-etal-2019-modeling, zou-lu-2019-text2math, ijcai2019-736}. However, different forms of equations can be formed to solve the same math problem, which often makes models fail. To resolve the equation duplication issues, various equation normalization methods are proposed in %\cite{wang-etal-2018-translating}
\cite{wang-etal-2018-translating,Wang2019TemplateBasedMW}
to generate a unique expression tree with the cost of losing the understanding of problem-solving steps in equation expressions.
% Inspired by the attention mechanism,
% % \cite{li-etal-2019-modeling}
% \cite{li-etal-2018-multi-head,li-etal-2019-modeling}
% solves the math word problems by splitting the problem text into question and quantity parts with multiple types of
% attention vectors using the Seq2Seq model.
% (the limited the source text to give problem description first and then pose question)
% To track the semantic meaning of operators and operands in expression, \cite{chiang-chen-2019-semantically} employ stack actions in their decoder to generate the expression in a postfix manner.
In contrast, we propose to use a Graph2Tree model to solve this task without any special mechanisms like equation normalization. To the best of our knowledge, this is the first work to use GNN to build a math word problem solver.

\section{Problem Formulation and Structure Object Construction}
\label{sec:graph}

\subsection{Graph-to-Tree Translation Task}
In this work, we consider the problem of translating a graph input to a tree output. In particular, we consider two important tasks - Semantic Parsing and Math Word Problem. Formally, we define both tasks as follows. The input side contains a set of text sequences, denoted as $S=\{s_1, s_2,\ldots, s_n\} \in \mathcal{S}$ where $s_i$ is a text sequence consisting of a sequence of word embeddings $s_i = \{w_1, w_2,\ldots, w_{|s_i|}\} \in \mathcal{W}$, where $\mathcal{W}$ is a pretrained word embedding space. We then construct a heterogeneous graph input $G = (V,E) \in \mathcal{G}$, where $V = [V_1 \ V_2]$ contains all of the original word nodes $V_1 \in \mathcal{V}_1$ as well as the relationship nodes $V_2 \in \mathcal{V}_2$ from the relationships of a parsing tree (i.e. dependency or constituency tree), and $E \in \mathcal{E}$ denotes if the two nodes are connected or not. The aim is to translate a set of heterogeneous graph inputs $G = \{g_1, g_2, \ldots, g_n\}$ into a set of tree outputs $T = \{t_1,t_2,...t_n\} \in \mathcal{T}$ where $t_i$ is a logic form or math equation consisting of a sequence of tree node token $t_i = \{y_1, y_2,\ldots, y_{|t_i|}\} \in \mathcal{Y}$. 
%In this work, we focus on the problem setting that we have a set of paired source graphs and target trees to learn the translator. For a more challenging problem setting that such an alignment is lacking, we leave these tasks as future work.

\subsection{Constructing Graph Inputs and Tree Outputs}
% \subsection{Constructing Syntactic Input Graphs}
To apply GNNs, the first step is to construct a graph input by combining the word sequence with their corresponding hidden structure information. How to construct such graphs is critical to incorporate the structured information and influences the final performance. Similarly, how to construct the tree outputs from logic form or math equations also play an important role in the final performance and model interpretability. In this section, we will introduce two methods for graph construction and one method for tree construction.

% \subsection{Combining Word Sequence with Dependency Parse Tree}

\begin{figure}[ht!]
  \centering
\includegraphics[width=6.5cm]{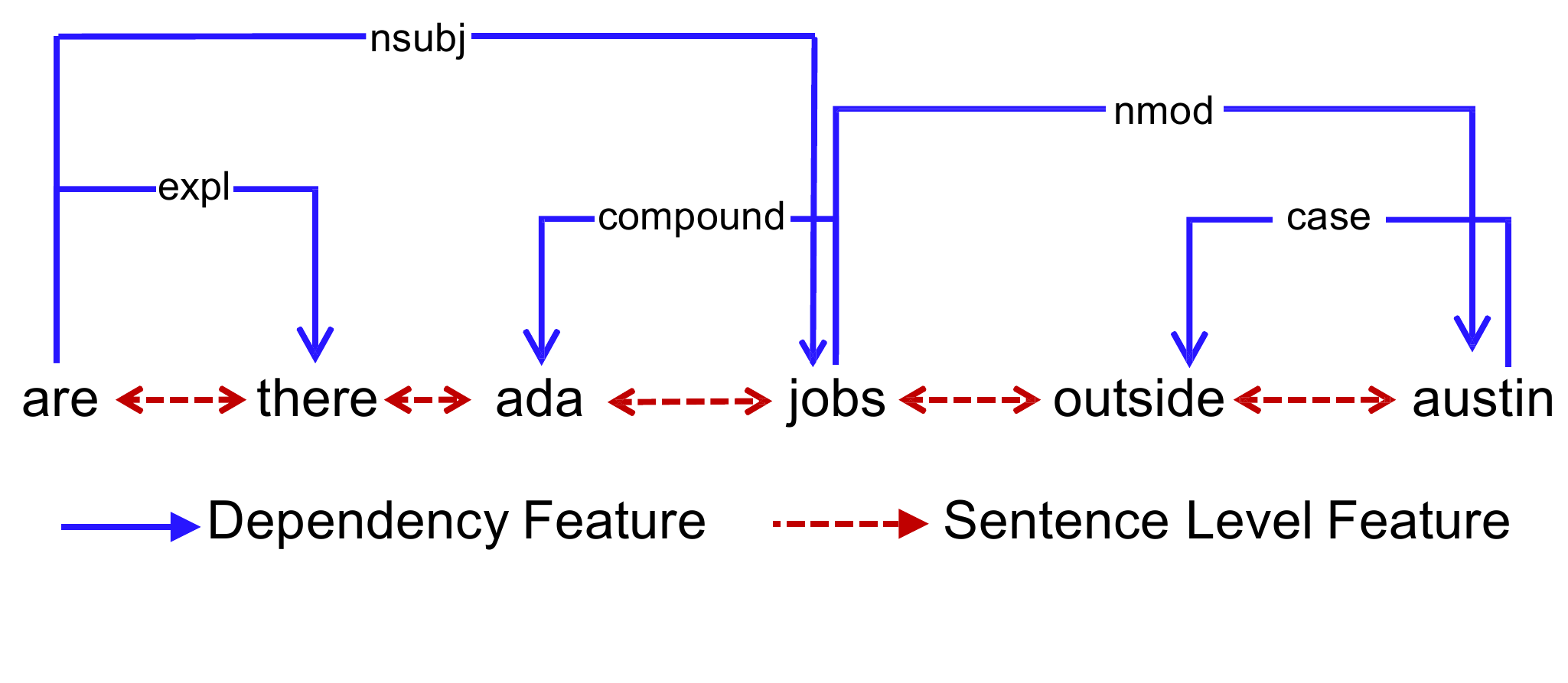}
 \vspace{-6mm}
\caption{Dependency tree augmented text graph}
  \label{fig:word-dependency}
%  \vspace{-2mm}
\end{figure}

\noindent\textbf{Combining Word Sequence with Dependency Parse Tree.}
The dependency parse tree not only represents various grammatical relationships between pairs of text words, but also
is shown to have an important role in transforming texts into logical forms \cite{reddy-etal-2016-transforming}. Therefore, the first method integrates two types of features by adding dependency linkages between corresponding word pairs in word sequence. Concretely, we transform a dependency label into a node, which is linked respectively with two word nodes with dependency relationship. Figure \ref{fig:word-dependency} gives such an example of constructed heterogeneous graph from a text.

% \subsection{Combining Word Sequence with Constituency Tree}
\begin{figure}[ht!]
  \centering
\includegraphics[width=8.0cm]{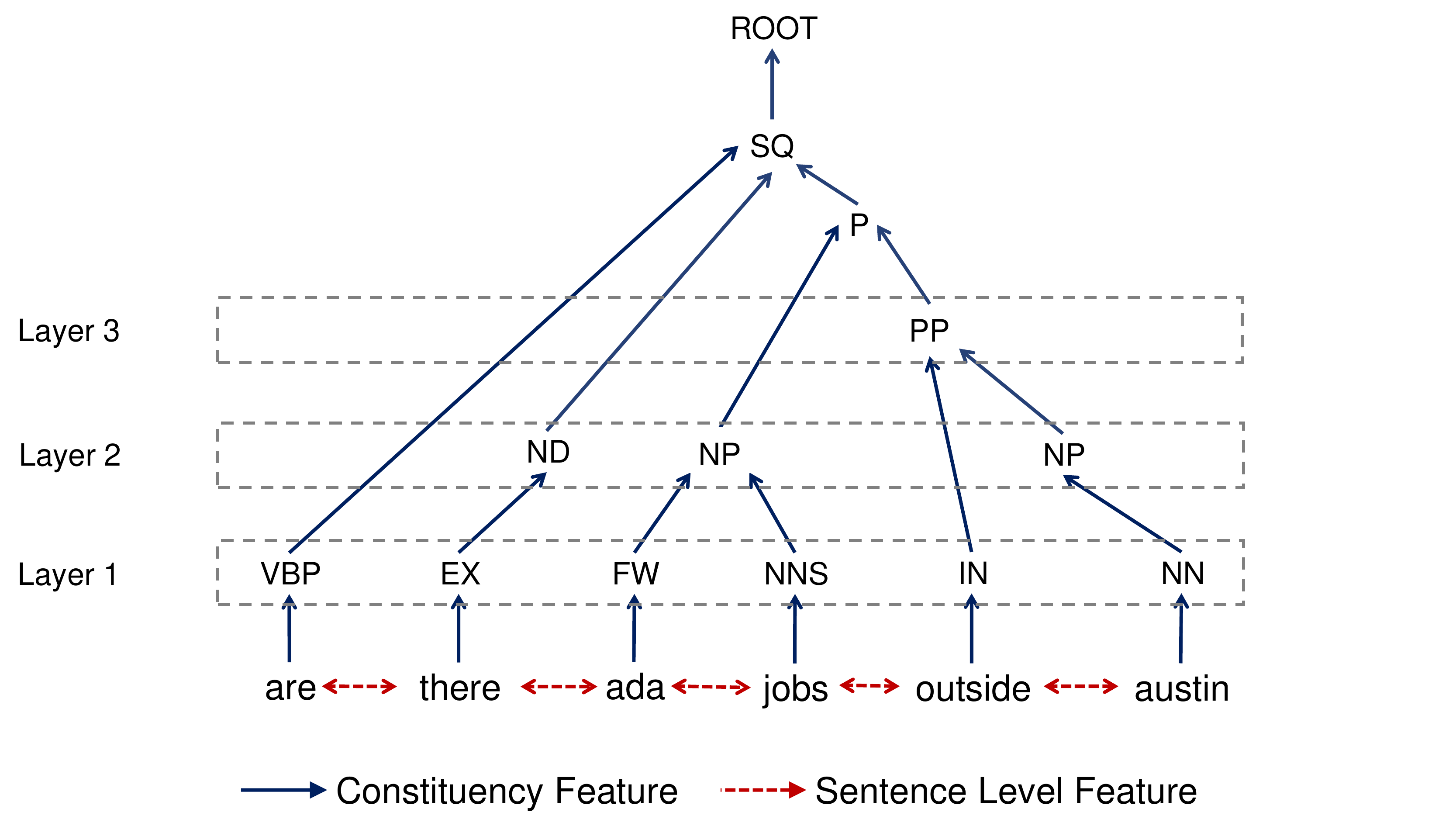}
 \vspace{-6mm}
\caption{Constituency tree augmented text graph}
  \label{fig:word-Constituency}
%  \vspace{-2mm}
\end{figure}

\noindent\textbf{Combining Word Sequence with Constituency Tree.}
The constituency tree contains the phrase structure information which is also critical to describe the word relationships and has shown to provide useful information for translation \cite{gu-etal-2018-top}. Since the leaf nodes in the constituency tree are the word nodes in the text, this method merges these nodes with the identical ones in the bi-directional word sequence chain to create the syntactic graph. Figure \ref{fig:word-Constituency} shows an example of constructed heterogeneous graph input.

\begin{figure}[ht!]
  \centering
\includegraphics[scale=0.25]{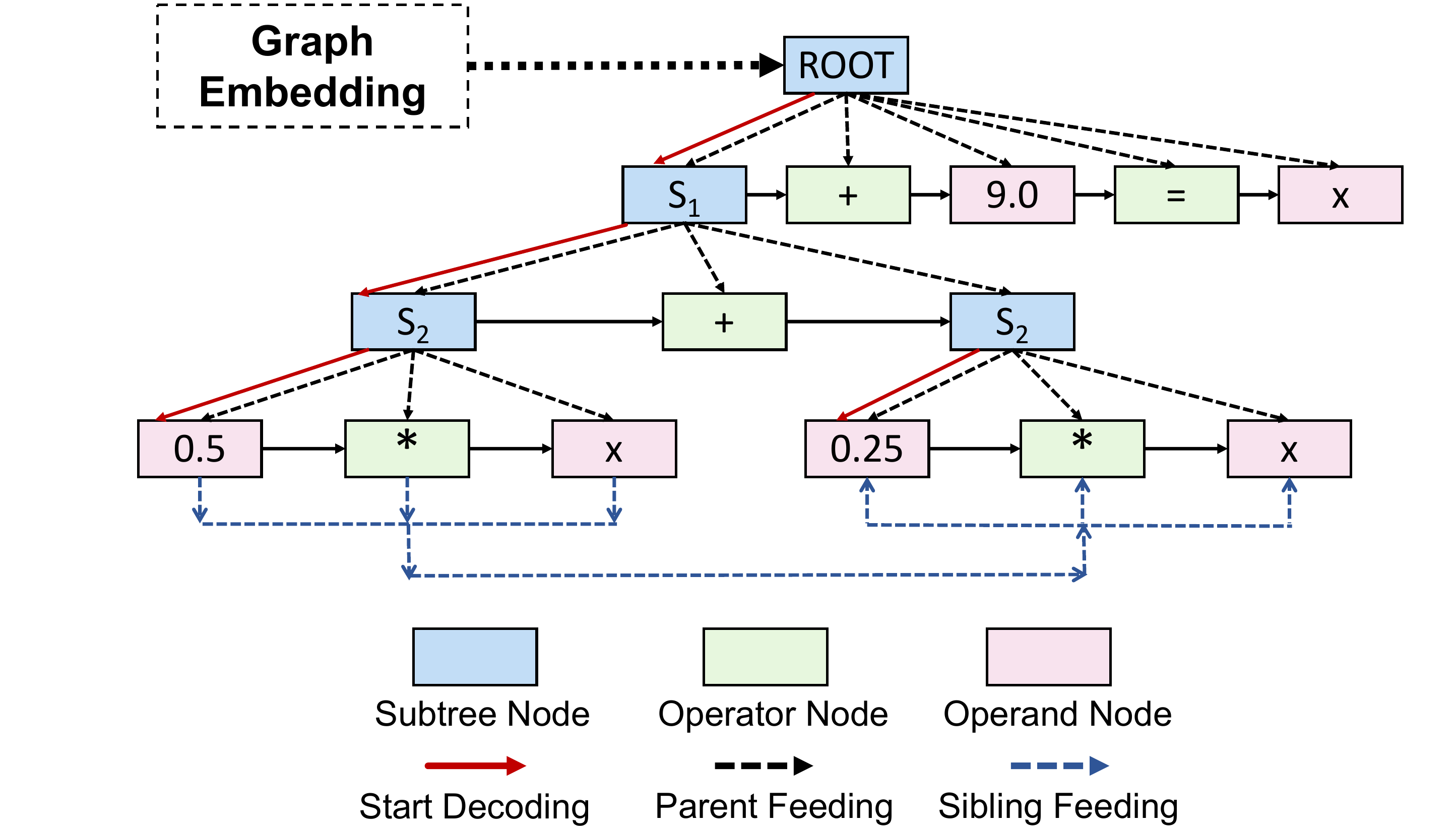}
% \vspace{0mm}
\caption{A sample tree output in our decoding process from expression "\textbf{( ( 0.5 * x ) + ( 0.25 * x ) ) + 9.0 = x}"}
  \label{fig:MWP_tree}
% \vspace{-2mm}
\end{figure}

\noindent\textbf{Constructing Tree Outputs.}
To effectively learn the compositional nature of our structured outputs, we need to firstly transform original outputs from logic forms or math equations to tree structured objects. Specifically, we follow the tree construction method in \cite{dong-lapata-2016-language}, which is a top-down manner to generate tree-structured outputs. In original outputs containing structural meaning symbols like brackets, we first extract sub-tree structures and replace these sub-tree structures with sub-tree symbols. Then we grow branches from the generated sub-tree symbols until all hierarchical structures in the original sequence are processed. Figure \ref{fig:MWP_tree} provides an example of constructed tree objects from mathematical expression.

\begin{figure*}
  \centering
%   \vspace{-2mm}
\includegraphics[width=15cm]{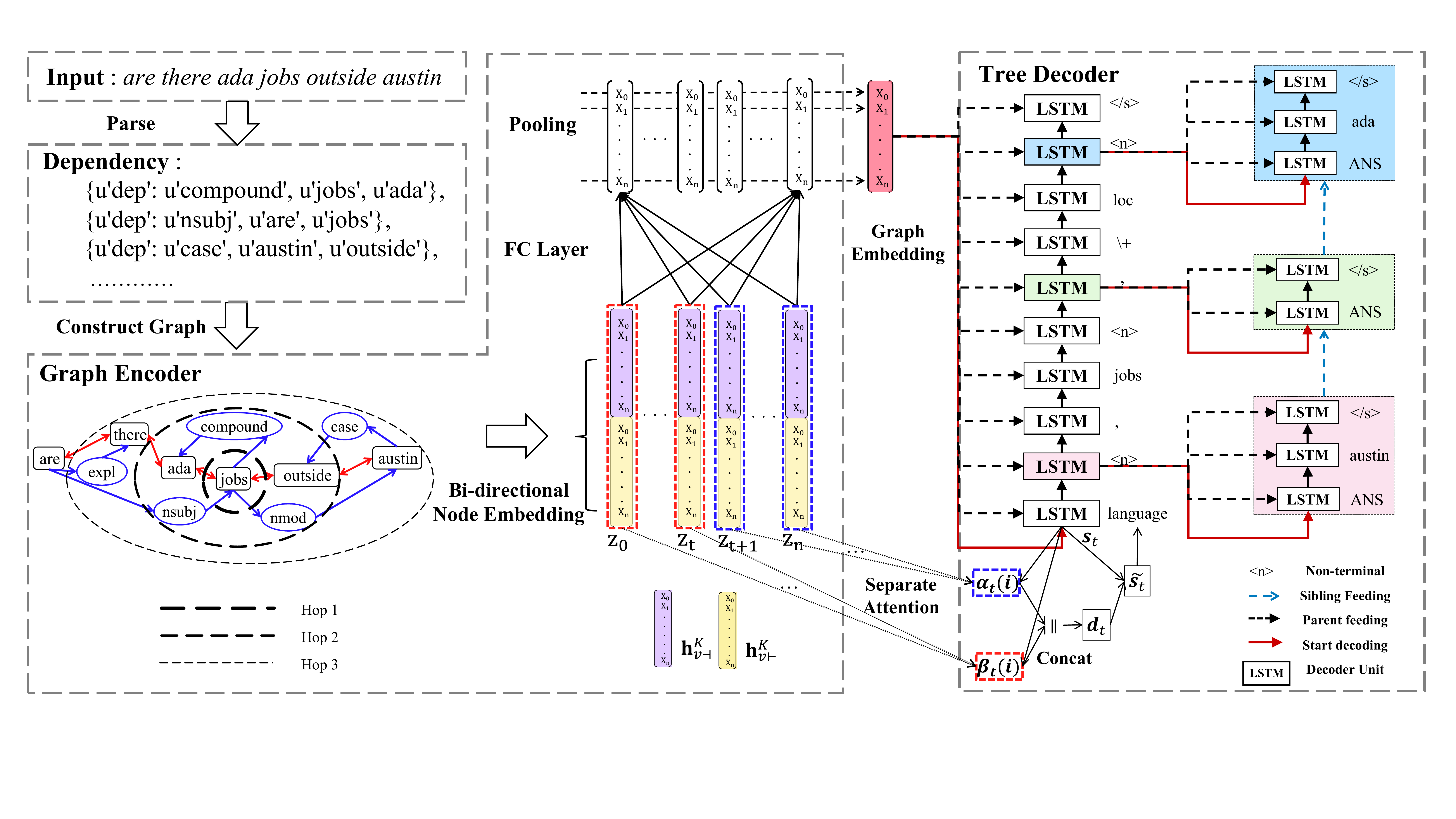}
 \vspace{-12mm}
\caption{Overall architecture of our Graph2Tree model. We use semantic parsing task as an example.}
  \label{fig:graph2tree}
%  \vspace{-2mm}
\end{figure*}

%parser
\section{Graph2Tree Neural Networks}
\label{sec:Graph2Tree Model}

We aim to learn a mapping that translates a heterogeneous graph-structured input $G$ and its corresponding tree-structured outputs $T$. 
%We design a novel Graph2Tree model, which consists of a \emph{graph encoder} that encodes a heterogeneous graph input into a vector representation and a \emph{tree decoder} that learns to generate tree output conditioned on the encoded graph-level representation. To capture the intuition of the modular translation process, the tree decoder also employs a separated attention mechanism to locate the corresponding source sub-graph when expanding the non-terminal. 
We illustrate the workflow of our proposed Graph2Tree model for semantic parsing in Figure \ref{fig:graph2tree}, and present each component of the model as follows.

% \subsection{Graph Encoder}
% In addition to the original sequence structure, their corresponding syntactic relations can be naturally incorporated into the input sequence which formulates an expressive graph data structure.
% As a result, we seek to exploit a graph neural network \cite{kipf2016semi} to effectively learn high-quality vector representation from graph input.
% . Each directional node embedding learns its vector representation by aggregating information from a node local neighborhood within $K$ hops of the graph.

% that is an inductive node embedding algorithm.

% \noindent \textbf{Graph Encoder.}
\subsection{Graph Encoder}
To effectively learn graph representations from our constructed heterogeneous text graph, we present a novel bidirectional graph node embeddings method - BiGraphSAGE. The proposed BiGraphSAGE extends the widely used GraphSAGE \cite{hamilton2017inductive} by learning forward and backward node embeddings of a graph $G$ in an interleaved fashion.
%The graph encoder \cite{xu2018graph2seq} to generate the bi-directional node embeddings (forward and backward node embeddings of a graph $G$).

In particular, consider a word node $v \in \mathcal{V}_1$ with pretrained word embedding \textbf{w}$_{v}$ like GloVe \cite{pennington-etal-2014-glove} as $v$'s initial attributes. We then generate the contextualized node embeddings \textbf{a}$_{v}$ for all nodes $v \in \mathcal{V}_1$ using Bi-directional Long Short Term Memory (BiLSTM) \cite{graves2013speech}. For a relationship node $v \in \mathcal{V}_2$, we initialize \textbf{a}$_{v}$ with randomized embeddings. These feature vectors are used as initial node embeddings $\textbf{h}_{v}^{0} = \textbf{a}_{v}$. Then each node embedding learns its vector representation by aggregating information from a node local neighborhood within $K$ hops of the graph.

\vspace{-5mm}
\begin{small}
\begin{gather}
% \tiny
% \textbf{a}_{v} = \text{BiLSTM}(\textbf{w}_{v}), \forall v \in \mathcal{V}_1 \ \text{or} \ \text{BiLSTM}(\textbf{r}_{v}), \forall v \in \mathcal{V}_2\\
% \textbf{a}_{v} = \text{BiLSTM}(\textbf{w}_{v}), \forall v \in \mathcal{V}_1\\
% \textbf{a}_{v} = \text{BiLSTM}(\textbf{r}_{v}), \forall v \in \mathcal{V}_2\\
% \textbf{h}_{v\vdash}^{0} = \textbf{a}_{v}, \textbf{h}_{\vdash v}^{0} = \textbf{a}_{v}, \forall v \in \mathcal{V}_1 \ \text{or} \ \mathcal{V}_2 \\
\textbf{h}_{\mathcal{N}_{\vdash}(v)}^{k}   = \textbf{M}_{\vdash}^{k}(\{\small \textbf{h}_{u\vdash}^{k-1}, \forall u \in \mathcal{N}_{\vdash}(v)\}) \\
% \textbf{h}_{v\vdash}^{k} = \sigma ( \textbf{W}^{k}\cdot \texttt{\small CONCAT}(\textbf{h}_{v\vdash}^{k-1}, \textbf{h}_{\mathcal{N}_{\vdash}(v)}^{k})) \\
\textbf{h}_{\mathcal{N}_{\dashv}(v)}^{k}   = \textbf{M}_{\dashv}^{k}(\{\small \textbf{h}_{u\dashv}^{k-1}, \forall u \in \mathcal{N}_{\dashv}(v)\})
% \textbf{h}_{v\dashv}^{k} = \sigma ( \textbf{W}^{k}\cdot \texttt{\small CONCAT}(\textbf{h}_{v\dashv}^{k-1}, \textbf{h}_{\mathcal{N}_{\dashv}(v)}^{k}))
\end{gather}
\end{small}
\vspace{-7mm}

\noindent where $k\in\{1,...,K\}$ is the iteration index and $\mathcal{N}$ is the neighborhood function of node $v$. \textbf{M}$_{\vdash}^{k}$ and \textbf{M}$_{\dashv}^{k}$ are the forward and backward aggregator functions. Node $v$'s forward (backward) representation \textbf{h}$_{v\vdash}^{k}$ (\textbf{h}$_{v\dashv}^{k}$) aggregates the information of nodes in $\mathcal{N}_{\vdash}(v)$ ($\mathcal{N}_{\dashv}(v)$).

Conceptually, one can choose to keep these node embeddings for each direction independently, which ignores interactions between two intermediate node embeddings during the training. Therefore, we fuse two intermediate unidirectional node embeddings at each hop as follows,

\vspace{-5mm}
\begin{small}
\begin{gather}
\textbf{h}_1 = \textbf{h}_{\mathcal{N}_{\vdash}(v)}^{k},  \
\textbf{h}_2 = \textbf{h}_{\mathcal{N}_{\dashv}(v)}^{k}  \\
\textbf{h}_{\mathcal{N}(v)}^{k} = \textbf{w}_g \odot \textbf{h}_1 + (1-\textbf{w}_g) \odot \textbf{h}_2, \\
\textbf{w}_g = \sigma(\vec{W}_{\!z} [\textbf{h}_1; \textbf{h}_2; \textbf{h}_1 \odot \textbf{h}_2;\textbf{h}_1-\textbf{h}_2])
\end{gather}
\end{small}
\vspace{-7mm}

\noindent where $\odot$ denotes component-wise multiplication, $\sigma$ is a sigmoid function and $\textbf{w}_g$ is a gating vector.

The graph encoder learns node embeddings $\textbf{h}_{v}^{k}$ by repeating the following process $K$ times:

\vspace{-4mm}
\begin{small}
\begin{gather}
\textbf{h}_{v}^{k} = \sigma ( \textbf{W}^{k}\cdot \texttt{\small CONCAT}(\textbf{h}_{v}^{k-1}, \textbf{h}_{\mathcal{N}(v)}^{k}))
\end{gather}
\end{small}
\vspace{-7mm}

\noindent where \textbf{W}$^{k}$ denotes weight matrices, $\sigma$ is a non-linearity function, $K$ is maximum number of hops.

The final bi-directional node embeddings $\textbf{z}_{v}$ is chosen to concatenate the two unidirectional node embeddings at the last hop,

\vspace{-4mm}
\begin{small}
\begin{gather}
\textbf{z}_{v} = \texttt{\small CONCAT}(\textbf{h}_{v\vdash}^{K}, \textbf{h}_{v\dashv}^{K}) \\
\textbf{g} = \texttt{\small MAXPOOL} (\texttt{\small FC}( \textbf{z} )).
\end{gather}
\end{small}
\vspace{-7mm}

\noindent After the bi-directional embeddings for all nodes $\textbf{z}$ are computed, we then feed the obtained node embeddings into a fully-connected neural network and apply the element-wise \textit{max}-pooling operation on all node embeddings to compute the graph-level vector representation $\textbf{g}$, where other alternative commutative operations such as \textit{mean} or attention based weighted \textit{sum} can be used as well.

\subsection{Tree Decoder}

We propose a new general tree decoder fully leveraging the outputs of our graph encoder, i.e. the bi-directional node embeddings and the graph embedding, and faithfully generating the tree-structured targets like logic forms or math equations.

Inspired by the thinking paradigm of human beings, our tree decoder at high level uses a divide-and-conquer strategy
splitting the whole decoding task into sub ones. Figure \ref{fig:MWP_tree} illustrates an example output of our tree decoder.
In this example, we firstly initialize the root tree node ${ROOT}$ with the graph embedding $\textbf{g}$, and then apply a sub-decoder on the ${ROOT}$ to generate a 1st-level coarse output containing a sub-tree node $S_1$. This $S_1$ is
further decoded with the similar sub-decoder to derive the 2nd-level coarse output. This procedure is repeated to
generate the 3rd-level output in which there is no sub-tree nodes. 
% This last level output of tree structure is the final output of our decoder.
In this way, we get the whole tree output in a top-down manner.

This whole procedure can be summarized as follows: 1) initialize the root tree node with the graph embedding from our
encoder and perform the first level decoding with our LSTM based sub-decoder; 2) for each newly generated sub-tree node,
a sub-decoder is applied to derive the next level coarse output; 3) repeat step 2 until there is no sub-tree nodes in
the last level of tree structure.

\subsubsection{Sub-Decoder Design}
In each of our sub-decoder task, the conditional probability of the generated word at step $t$ is calculated as follows:

\vspace{-1mm}
\begin{small}
\begin{equation}
	\label{sub-decoder}
  p(y_t|\textbf{y}_{<t}, \textbf{x}) = f_{predict}(\textbf{s}_{t})
\end{equation}
\end{small}
\vspace{-5mm}

\noindent where $\textbf{x}$ denotes vectors of all input words, $y_{t}$ is the predicted output word at $t$, $\textbf{s}_{t}$ is the decoder hidden
state at $t$, and $f_{predict}$ is a non-linear function.

The key component of Eq. \eqref{sub-decoder} is the computation of $\textbf{s}_{t}$. Conceptually, this value is calculated as $\textbf{s}_{t} = f_{decoder}(y_{t-1}, \textbf{s}_{t-1})$, where $f_{decoder}$ is usually a RNN unit. We propose two improvements on top of it, \textit{parent feeding} and \textit{sibling feeding}, to feed more information for decoding sub-tree nodes.

\textbf{Parent feeding}. For a sub-task in our tree decoding process, we aim to expand the sub-tree node in the parent layer. Therefore, it is reasonable to take the sub-tree node embedding $\textbf{st}_i$ into consideration. Therefore, we add the sub-tree node embedding as part of our input at every time-step, in order to capture the upper-layer information for decoding.

\textbf{Sibling feeding}. Besides the information from parent nodes, if two sub-tree nodes share the same parent node, then these two sub-tasks can also be related. Inspired by this observation, we employ the sibling feeding mechanism to feed the preceding sibling sentence embedding to the sub-task related to its closet neighbor sub-tree node. For example, imagine $p_1$ is the parent node of $c_1$, $c_2$, and we feed both embeddings of $p_1$ and $c_1$ when decoding $c_2$.

Therefore, our sub-decoder calculates the decoder hidden state $\textbf{s}_t$ as follows:

\vspace{-4mm}
\begin{small}
\begin{equation}
  \textbf{s}_t = f_{decoder}(y_{t-1}, \textbf{s}_{t-1};\textbf{st}_{parent};\textbf{st}_{sibling})
\end{equation}
\end{small}
\vspace{-6mm}

\noindent where $\textbf{st}_{parent}$ stands for sub-tree node embedding from parent layer and $\textbf{st}_{sibling}$ is the sentence embedding of
the closest preceding sibling. By fully utilizing the information from parent nodes and sibling nodes, our tree decoder can effectively generate target hierarchical outputs.

\subsection{Separate Attention Mechanism to Locate Source Sub-graph}

Various attention mechanisms have been proposed \cite{bahdanau+al-2014-nmt,luong-etal-2015-effective} to incorporate the hidden
vectors of the inputs into account during the decoding processing.
In particular, the context vector $\bm{{s_{t}}}$ depends on a set of bidirectional node representations of the source graph (\textbf{z$_{1}$},...,\textbf{z$_{|V|}$})
to which the decoder locates the source sub-graph. Since our graph input is essentially a heterogeneous graph with two
different input sources (word nodes with relationship nodes of a parsing tree), we propose to employ a separated attention mechanism over the node representations corresponding to the different node types:

\vspace{-4mm}
\begin{small}
\begin{equation}
\alpha_{t(v)} = \frac{\exp(score( {\textbf{z}_v},\textbf{s}_t))}{\exp(\sum_{k=1}^{V_1} score({\textbf{z}_{k}},\textbf{s}_t))}, \forall v \in \mathcal{V}_1
\end{equation}
\vspace{-2mm}
\begin{equation}
\beta_{t(v)} = \frac{\exp(score( {\textbf{z}_v},\textbf{s}_t))}{\exp(\sum_{k=1}^{V_2} score({\textbf{z}_{k}},\textbf{s}_t))}, \forall v \in \mathcal{V}_2
\end{equation}
\end{small}
\vspace{-2mm}

\noindent where the $score(\cdot)$ function estimates the similarity of $\textbf{z}_v$ and $\textbf{s}_t$.
Then, we compute the context vectors $\textbf{c}_{v1}$ and $\textbf{c}_{v2}$, respectively.

\vspace{-4mm}

\begin{small}
\begin{align}
\textbf{c}_{v_1} &= \sum \alpha_{t(v)} \textbf{z}_v, \forall v \in \mathcal{V}_1 \\
\textbf{c}_{v_2} &= \sum \beta_{t(v)} \textbf{z}_v, \forall v \in \mathcal{V}_2
\end{align}
\end{small}

\vspace{-6mm}

We concatenate the context vector $\textbf{c}_{v_1}$, context vector $\textbf{c}_{v_2}$ and decoder hidden state $\textbf{s}_t$ to compute the final attention hidden state at this time step as:
% $\tilde{s_t} = \tanh(W_c \cdot [c_u;c_v;s_t]+b)$,

\vspace{-1mm}
\begin{small}
\begin{equation}
    \tilde{\textbf{s}_t} = \tanh(W_c \cdot [\textbf{c}_{v_1};\textbf{c}_{v_2};\textbf{s}_t]+b_c)
\end{equation}
\end{small}
\vspace{-5mm}

\noindent where $W_c$ and $b_c$ are learnable parameters. The final context vector $\bm{\tilde{s_{t}}}$ is further used for decoding tree structured outputs. The output probability distribution over a vocabulary at the current time step is calculated by:

\vspace{-5mm}
\begin{small}
\begin{equation}
    p(y_t|y_1,y_2,\ldots,y_{t-1}, g)=softmax(W_v\tilde{\textbf{s}_t}+b_v)
\end{equation}
\end{small}
\vspace{-6mm}

\noindent where $W_v$ and $b_v$ are learnable parameters. Our model is then jointly trained to maximize the conditional log-probability of the target tree given a heterogeneous graph input $g$.

\section{Experiments}
\label{sec:exp}
In this section, we evaluate the effectiveness and generality of Graph2Tree model on two important tasks -- Semantic Parsing and Math Word Problem. The code and data for our Graph2Tree model are provided for research purpose \footnotemark[1].
% \footnotetext[1]{Code will be released upon the paper acceptance.}
\footnotetext[1]{\url{https://github.com/IBM/Graph2Tree}}

\subsection{Experiments for Semantic Parsing}
\textbf{Datasets.}
    We evaluate our Graph2Tree on three totally-different benchmark datasets, JOBS \cite{Zettlemoyer05learningto}, GEO \cite{Zettlemoyer05learningto}, and ATIS \cite{dahl-etal-1994-expanding}, for the semantic parsing task. The first one
    JOBS is a set of 640 queries from a job listing database, the second one GEO is a set of 880 queries on a database of
    U.S. geography, and the last one ATIS is a dataset of 5410 queries from a flight booking system. We utilize the same
    train/dev/test split standard as used in previous works. 
    % \cite{Zettlemoyer05learningto}. 
    We adopt the data preprocessing provided by~\cite{dong-lapata-2016-language}. Natural language utterances are in lower case and stemmed, and entity mentions are
    replaced by numbered markers. For the graph construction, we use the dependency parser and constituency parser from CoreNLP \cite{manning-etal-2014-stanford}.

\noindent\textbf{Settings.}
    We use the Adam optimizer \cite{kingma2014adam} with a batch size of 20. For the JOBS and GEO datasets, our
    hyper-parameters are cross-validated on the training sets. For ATIS, we tune them on the development set. The
    learning rate is set to 0.001. In graph encoder, the BiRNN we use is a one-layer BiLSTM with a hidden size of 150,
    and the hop size in GNN is chosen from \{2,3,4,5,6\}. The decoder we employ is a one-layer LSTM with a hidden size
    of 300. The dropout rate is chosen from \{0.1,0.3,0.5\}. 
    %We use the ReLU as our non-linear function and the greedy search as our inference strategy for decoding.
    %, and embedding matrix is initialized with GloVe vectors(300 dimension) from \citet{pennington-etal-2014-glove}.

\noindent\textbf{Baselines.} We compare our model against several state-of-the-art neural semantic parsers: i)
Seq2Seq model with a Copy mechanism \cite{jia-liang-2016-data}; ii) Seq2Seq and Seq2Tree models \cite{dong-lapata-2016-language}; iii)
Graph2Seq model \cite{xu2018graph2seq}. We report the exact-match accuracy for each baseline on all three benchmarks.

\begin{table}[!ht]
% \vspace{-2mm}
\centering
\small
% \caption{Exact-match accuracy comparison on all three benchmarks \textit{JOBS}, \textit{GEO}, and \textit{ATIS} for SP task}
\newcommand{\Bd}[1]{\textbf{#1}}
% \vspace{-2mm}
\begin{tabular}{|c||c|c|c|}
\hline
{\bf Methods} & {\bf \textit{JOBS}} & {\bf \textit{GEO}} & {\bf \textit{ATIS}}\\
\hline
Jia et al.(2016) & - & 85.0 & 76.3 \\
\hline
Dong et al.(2016)-Seq2Seq & 87.1 & 84.6 & 84.2\\
\hline
Dong et al.(2016)-Seq2Tree & 90.0 & 87.1 & \Bd{84.6} \\
\hline
% Rabinovich et al. (2017) & 92.9 & 85.7 & \Bd{85.3}\\
% \hline
% Xu and Wu (2018)-Graph2Seq\footnotemark[1] & 88.6 & 85.7  & 83.8\\
Xu et al.(2018)-Graph2Seq\footnotemark[2] & 88.6 & 85.7  & 83.3 \\
\hline
% {\bf Graph2Tree} & \Bd{92.9} & \Bd{88.9} & 84.6 \\
{\bf Graph2Tree} & \Bd{92.9} & \Bd{88.9} & \Bd{84.6} \\
\hline
% {\bf BASELINE2Seq} & 88.1 & 84.9 \\
% \hline
\end{tabular}
% \vspace{-2mm}
\caption{Exact-match accuracy comparison on all three benchmarks \textit{JOBS}, \textit{GEO}, and \textit{ATIS} for SP task}
\label{tab:compare}
\end{table}

\footnotetext[2]{We run our own implementation of Graph2Seq on thesedatasets using PyTorch.}

\begin{table}[ht!]
% \vspace{-4mm}
    \centering
    \scriptsize
    % \caption{Case study of SP input: ``\textit{what jobs can a delphi developer find in san antonio on
    % windows ?}''}
     \vspace{-0mm}
    \newcommand{\Bd}[1]{\textbf{#1}}
    \begin{tabular}{|c|c|}
    \hline
    \Bd{Methods} & \Bd{Translated logic form results} \\
    \hline
    \multirow{3}*{Reference str} & job (ANS), language (ANS, 'delphi'),\\
    ~ & title (ANS, 'developer'), loc (ANS, 'san antonio'),\\
    ~ & platform (ANS, 'windows')\\ \hline \hline

    \multirow{3}*{\textbf{Graph2tree}} & job (ANS), language (ANS, 'delphi'),\\
    ~ & title (ANS, 'developer'), loc (ANS, 'san antonio'),\\
    ~ & platform (ANS, 'windows' )\\ \hline

    \multirow{2}*{Graph2seq} & job (ANS), language (ANS, 'delphi'),\\
    ~ & title (ANS, 'developer'), platform (ANS, 'windows')\\ \hline

    \multirow{2}*{Seq2seq} & job (ANS), language (ANS, 'delphi'),\\
    ~ & title (ANS, 'developer'), loc (ANS, 'san antonio')\\ \hline

    % \hline
    % Graph2tree & \multicolumn{1}{|m{5cm}|}{job ( ANS ) , language ( ANS , 'delphi' ) , title ( ANS , 'developer' ) , loc ( ANS , 'san antonio' ) , platform ( ANS , 'windows' )} \\\hline
    % Graph2seq & job ( ANS ) , language ( ANS , 'delphi' ) , title ( ANS , 'developer' ) ,
    % \\&platform ( ANS , 'windows' )
    % \\ \hline
    % Seq2seq & job ( ANS ) , language ( ANS , 'delphi' ) , title ( ANS , 'developer' ) ,
    % \\&loc ( ANS , 'san antonio' )
    \hline
    \end{tabular}
    % \vspace{-2mm}
    \caption{\label{tab:res} Case study of SP input: ``\textit{what jobs can a delphi developer find in san antonio on
    windows ?}''}
    \end{table}

% \begin{figure}[ht!]
% % \vspace{-1mm}
%   \centering
% \includegraphics[width=7.5cm]{Graphs/dep_example_2.pdf}
% %  \vspace{-6mm}
% \caption{Graph with word sequence and dependency tree}
%   \label{fig:dep tree}
%  \vspace{-6mm}
%\caption{Graph with word sequence and dependency tree for input sentence ``\textit{what jobs can a delphi developer
%find in san antonio on windows ?}''}
%  \label{fig:dep tree}
% \vspace{-6mm}
% \end{figure}

% \subsubsection{Results}
\noindent\textbf{Results.} Table \ref{tab:compare} shows that our proposed Graph2Tree outperforms or achieves comparable exact-match accuracy
compared to other state-of-the-art baselines, highlighting the effectiveness of our proposed model by exploiting full utilization of structural information in both inputs and outputs.

\noindent\textbf{Case study}. Next we analyze the different decoding results of all models for an example case in Table \ref{tab:res}.
The challenge in semantic parsing is the high-order neighborhood estimation of the noun key word ``jobs" to its attribute words
``windows" and ``san antonio". It is hard for the traditional sequence encoder to encode high-order neighborhood (long-range dependency). For instance, there are 10 hops between the word ``jobs" and
``windows" according to the sequential dependency, while there are only two hops if we introduce the syntactic dependency
information. Therefore, syntactic graph with graph encoder is an effective way to learn a high-quality representation
for decoding. This partially explains why our Graph2tree model outperforms Seq2Seq and Seq2Tree models.
% The results in table \ref{tab:res} confirm our analysis.

\begin{table}[ht!]
\centering
% \vspace{mm}
% \scriptsize
% \caption{Ablation study of Graph2Tree on the semantic parsing (JOBS and GEO) $\&$ math word problem (MAWAPS). We employ exact match accuracy for SP and solution accuracy for MWP.}
\vspace{-0mm}
\newcommand{\Bd}[1]{\textbf{#1}}
\begin{tabular}{|l||c|c|}
\hline
\Bd{Methods} & \Bd{JOBS} & \Bd{GEO} \\
\hline
Full model & 92.9 & 88.9\\\hline
w/o const tree & 90.0 & 86.8\\
\hline
w/ original GraphSage & 90.7 & 88.2  \\ \hline
w/ only parent feeding & 91.4 & 87.9 \\ \hline
w/ only sibling feeding & 89.2 & 84.3  \\ \hline
w/o parent \& sibling feeding & 88.6 & 83.9 \\ \hline

w/o separated attention & 83.6 & 77.1 \\
\multicolumn{1}{|c||}{w/ uniform attention} & 90.7 & 87.1 \\
\hline
w/o bilstm& 89.3 & 86.4 \\
\hline
\end{tabular}
\vspace{-0mm}
\caption{\label{tab:ablation study} Ablation study of Graph2Tree on the semantic parsing (JOBS and GEO). We employ exact match accuracy as evaluation metric. }
%\textit{Full model}: our best settings for Graph2Tree with our constituency graph construction method. \textit{w/o const tree}: use only word nodes in original sentences as our input. \textit{w/ original GraphSage}: replace our BiGraphSAGE graph encoder with original GraphSage. \textit{w/o separated attention}: do not use attention mechanism. \textit{w/ uniform attention}: replace separate attention with vanilla attention mechanism. \textit{w/o bilstm}: do not use Bi-LSTM in our encoder. For \textit{w/o parent and sibling feeding} and \textit{w/ only sibling feeding}: it should be noted that we just do not feed parent hidden state for all children node decoding steps, but for the first descendant decoding step, we need to do parent feeding to keep the information flow complete, otherwise the decoding process will fail.} 
\end{table}

\noindent\textbf{Ablation study.}
Table~\ref{tab:ablation study} presents the ablation study on our Graph2Tree using a constituency tree based graph (on SP datasets JOBS and GEO). This is done with test sets (JOBS and GEO have no dev set). Firstly, we observe that the syntactic information in the constituency tree, which is helpful for describing word relationships, is critical to our overall performance.
And we found that our bidirectional GraphSAGE, encoding from both forward and backward nodes according to edge direction, is proved to enhance the final performance.
Furthermore, parent feeding and sibling feeding mechanism, which can enrich both the paternal and fraternal information in decoding, also play important roles in the whole model. 
In addition, designed for different types of nodes in the input graph, the separate attention mechanism is proved useful in our model.
Last but not least, it is also necessary to use Bi-LSTM in the encoder to learn the contextualized word embeddings from the word sequences.

\subsection{Experiments for Math Word Problems}
\noindent\textbf{Datasets.}
% We here evaluate our Graph2Tree model on Math Word Problems (MWPs) for automatically translating the text descriptions
% into the math equations.
We here evaluate our Graph2Tree model on two benchmark datasets, MAWPS \cite{koncel-kedziorski-etal-2016-mawps} and MATHQA \cite{amini-etal-2019-mathqa}, for the Math Word Problems automatically solving task.
% The MAWPS dataset is proposed by  as a standard benchmark for
% testing different math word problem solvers. 
The MAWPS dataset is a Math Word Problem dataset in English and
contains 2373 pairs after harvesting equations with single unknown variable. 
% We follow the same train/dev/test split as \cite{robaidek2018data}.
The other MATHQA dataset is a recently proposed large-scale Math Word Problem dataset with 37k English pairs, where each math expression is corresponding to an annotated formula for better interpretability. This dataset is more difficult for covering complex multivariate problems. 
% The train/dev/test split is 29837/4475/2985.
% It also provides tools to tune the lexical and template overlap of datasets and interfaces for expanding dataset from web-sourced data.

%\noindent\textbf{Settings.}
% By following \cite{robaidek2018data}, we randomly split the original dataset with the same train/dev/test split ratio
% and perform the pre-processing - transform natural language sentences to lower cases and use number mappings to replace
% the numbers with indexed markers.
%We first mapped the numbers in questions to indexed markers.
%And we employ GloVe pretrained embeddings with a dimension size of 300. Our batch size is set to 30.
% We use Adam with learning rate of 0.001 for optimization.
%For other hyperparameter and graph construction settings, we employ a similar scheme in the semantic parsing subsection.
% In our graph encoder, we use a one-layer BiLSTM
% with hidden layer size of 150 to encode word order features in the problem description sentences. The hop size of GNN is
% chosen from \{2,3,4,5,6,7\}. For the decoder, we use a one-layer LSTM with hidden size of 300. Dropout rate is chosen from \{0.1,0.3\}. %We employ greedy search in our inference process and evaluate with solution accuracy.

% For graph construction, we use similar construction scheme in the semantic parsing subsection.

\begin{figure*}[!ht]
\centering

\subfigure[A graph-to-tree translation example]{
\centering
\includegraphics[width=5.4cm]{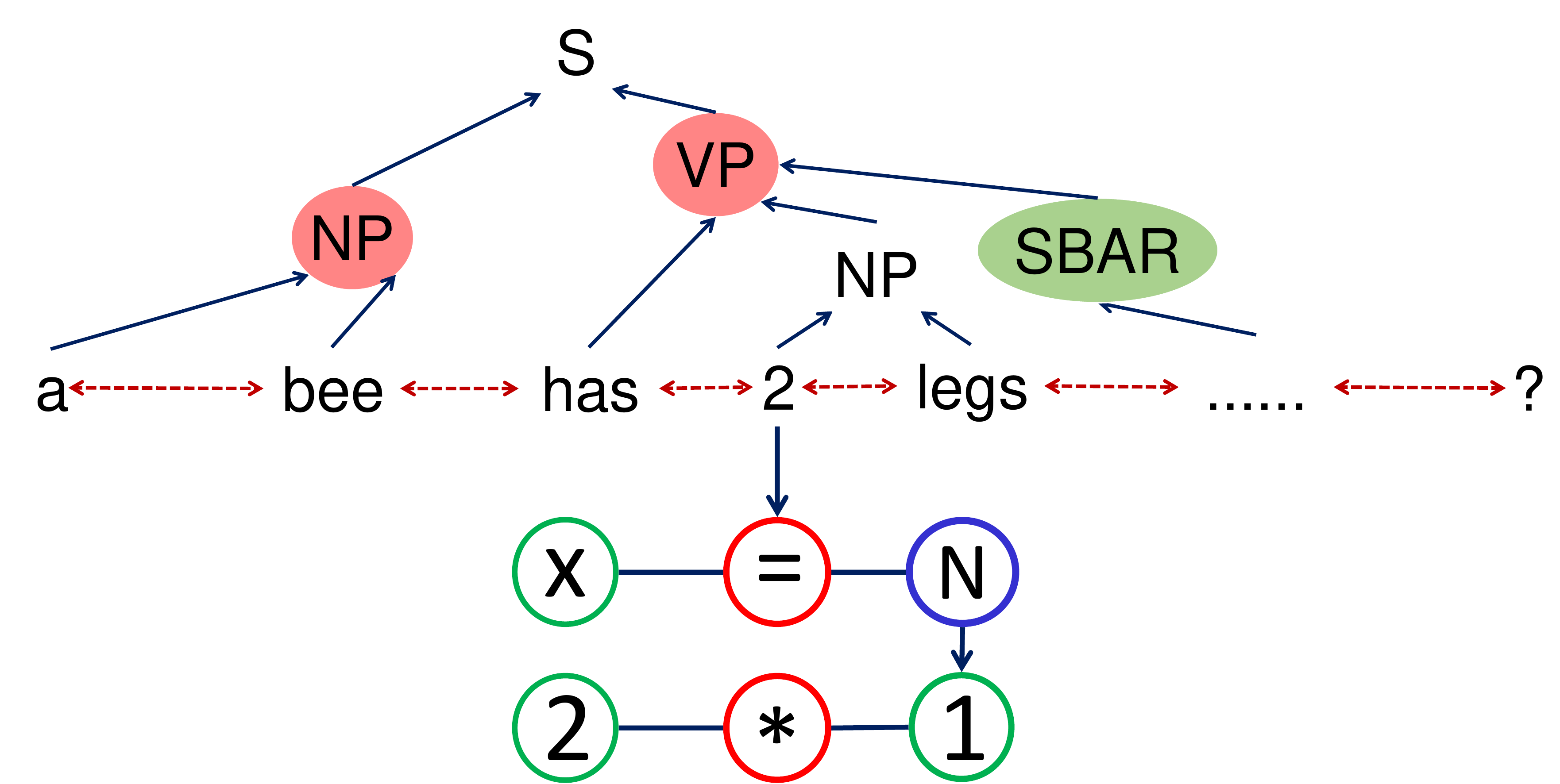}
\label{fig:att1}
}
\subfigure[Attention for word nodes ]{
\centering
\includegraphics[width=5.8cm]{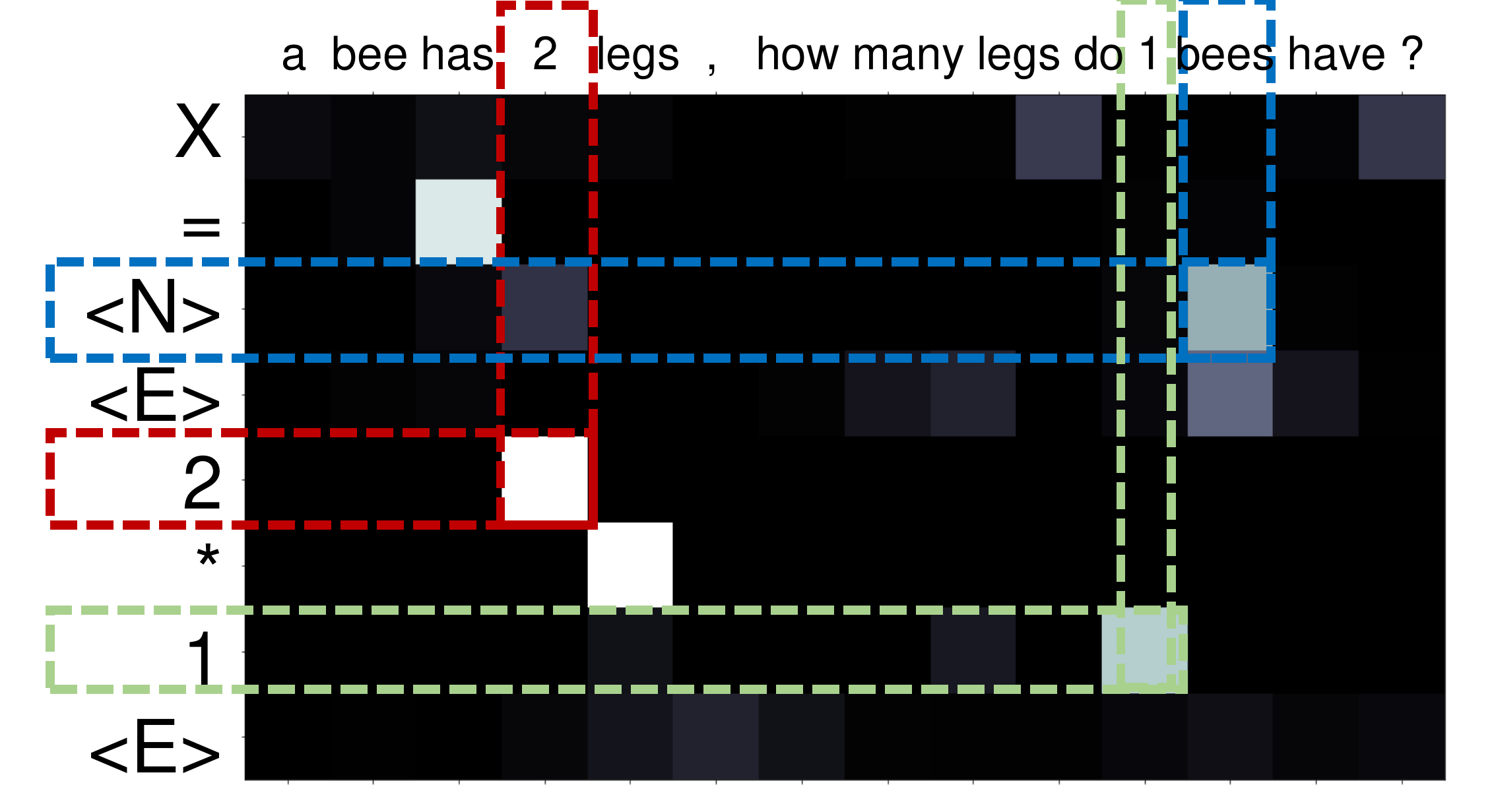}
\label{fig:att2}
}
\subfigure[Attention for structure nodes]{
\centering
\includegraphics[width=3.8cm]{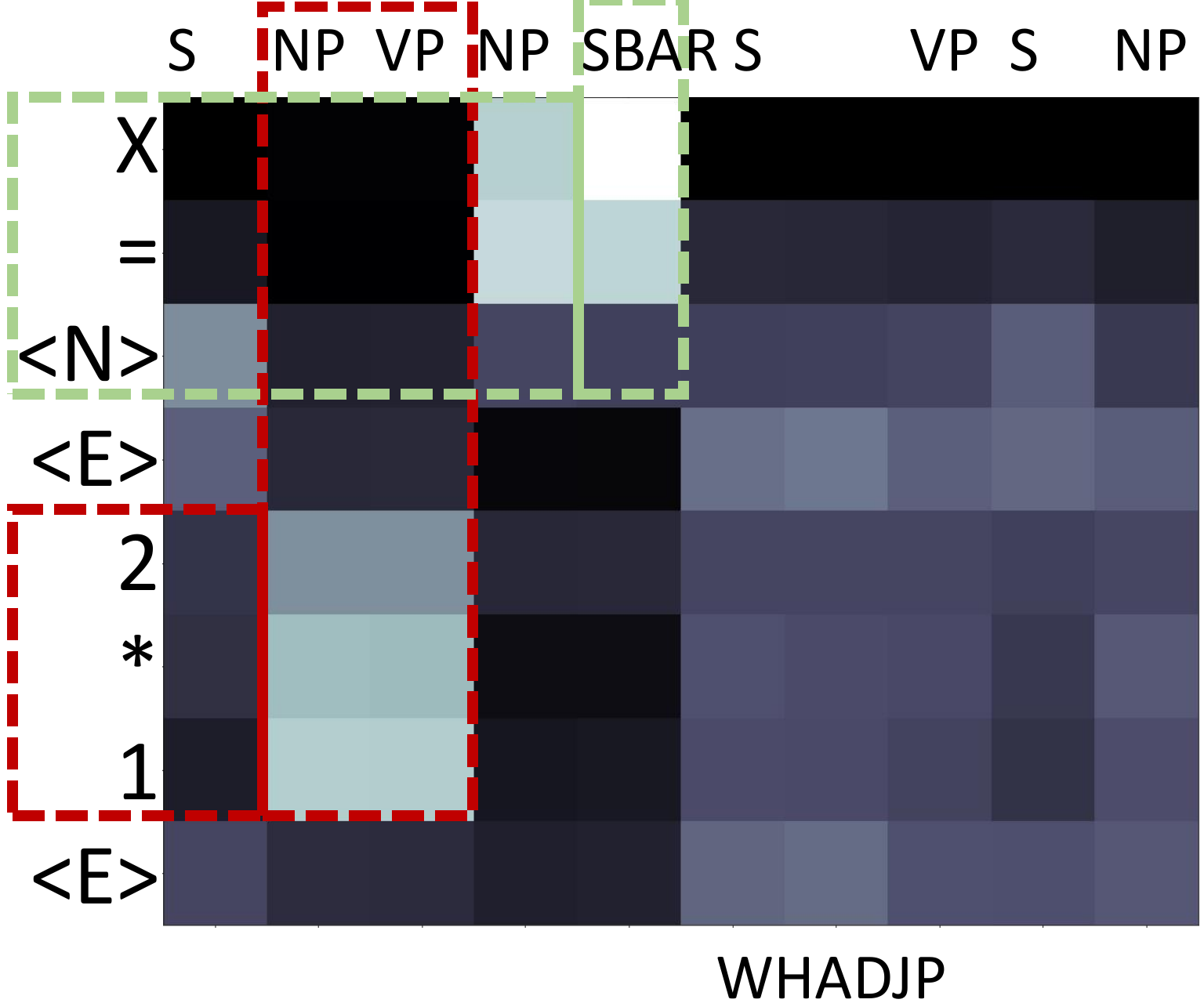}
\label{fig:att3}
}
% \vspace{-3mm}
\caption{Effect visualization of our separated attentions on both word and structure nodes in a graph.}
% \vspace{-4mm}

\label{fig:att_vis}
\end{figure*}

\noindent\textbf{Baselines.} We compare our Graph2Tree model against several state-of-the-art methods. We report the solution accuracy for each baseline in test set.
On MAWPS, our baselines are: i) Retrieval, Classification, and Seq2Seq \cite{robaidek2018data}; ii) Seq2Tree \cite{dong-lapata-2016-language}; iii) Graph2Seq \cite{xu2018graph2seq}; iv) MathDQN \cite{wang2018mathdqn}; v) T-RNN \cite{Wang2019TemplateBasedMW}; vi) Group-Att \cite{li-etal-2019-modeling}. 
On MATHQA, our baselines are: i) Sequence-to-program \cite{amini-etal-2019-mathqa}; ii) TP-N2F \cite{chen2019natural}; iii) Seq2Seq, Seq2Tree and Graph2Seq.
%We compare our Graph2Tree model with recent state-of-the-art methods for solving math word problem. The first one relies on classification models to choose proper equation templates according to problem descriptions and then fills generated templates with numeric operands. And the other one employs an encoder-decoder framework, usually a seq2seq model, to generate mathematical expressions in an end-to-end way with less human intervention.

\begin{table}[!ht]
\small
\centering
% \vspace{-2mm}
% \caption{Solution accuracy comparison on \textit{MAWPS}}
\newcommand{\Bd}[1]{\textbf{#1}}
\vspace{-0mm}
\begin{tabular}{|c|c|c|}
\hline
\multicolumn{2}{|c|}{\bf Methods} & {\bf MAWPS}\\
\hline
\multicolumn{2}{|c|}{Oracle} & 84.8 \\
\hline \hline
\multirow{2}*{Retrieval} & Jaccard & 45.6 \\
~ & Cosine & 38.8 \\
\hline
\multirow{2}*{Classification} & BiLSTM & 62.8 \\
~ & Self-attention & 60.4 \\
\hline
\multirow{2}*{Seq2seq} & LSTM & 25.6 \\
~ & CNN & 44.0 \\
% ~ & DNS & 59.9 \\
\hline
\multicolumn{2}{|c|}{Seq2Tree} & 65.2\\
\hline
\multicolumn{2}{|c|}{Graph2Seq} & 70.4 \\
\hline
\multicolumn{2}{|c|}{MathDQN} & 60.25 \\
\hline
\multirow{3}*{T-RNN} & Full model & 66.8 \\
~ & W/o equantion normalization & 63.9\\
~ & W/o self-attention & 66.3\\
\hline
\multicolumn{2}{|c|}{Group-Att} & 76.1 \\
\hline
\multirow{2}*{\bf Graph2Tree} & with constituency graph & \textbf{78.8} \\
~ & with dependency graph & {76.8} \\
\hline
\end{tabular}
\vspace{-0mm}
\caption{\label{tab:mwp_acc} Solution accuracy comparison on \textit{MAWPS}}
\end{table}

\begin{table}[!ht]
\small
\centering
\newcommand{\Bd}[1]{\textbf{#1}}
\vspace{-0mm}
\begin{tabular}{|c|c|c|}
\hline
\multicolumn{2}{|c|}{\bf Methods} & {\bf MATHQA}\\
\hline
\multicolumn{2}{|c|}{Seq2Prog} & 51.9 \\
\hline
\multicolumn{2}{|c|}{Seq2Prog+Cat} & 54.2 \\
\hline
\multicolumn{2}{|c|}{TP-N2F} & 55.95 \\
\hline
\hline
\multicolumn{2}{|c|}{Seq2seq} & 58.36 \\
\hline
\multicolumn{2}{|c|}{Seq2Tree} & 64.15\\
\hline
\multicolumn{2}{|c|}{Graph2Seq} & 65.36 \\
\hline
\multirow{2}*{\bf Graph2Tree} & with constituency graph & \textbf{69.65} \\
~ & with dependency graph & {65.66} \\
\hline
\end{tabular}
\vspace{-0mm}
\caption{\label{tab:mathqa_acc} Solution accuracy comparison on \textit{MATHQA}}
\vspace{-4mm}
\end{table}

\noindent\textbf{Results.}
As shown in Table \ref{tab:mwp_acc}, our Graph2Tree model consistently outperforms other state-of-the-art baselines by a large margin up to 10 points absolute accuracy except Group-Att baseline. To the best of our knowledge, we make the first attempt to employ the graph neural network for solving Math Word Problems, and our Graph2Tree model with constituency graph achieves the best performance so far on this MAWPS benchmark. We have observed similar conclusions on a more challenging and larger dataset -- MATHQA. This highlights the importance of having our Graph2Tree neural networks that can leverage the structured information from both inputs and outputs for automatic solving of math problems.

%Compared to the methods mentioned above that often utilize Seq2Seq or some special mechanisms such as Equation Normalization (EN) method, we make the first attempt to employ the graph neural network for solving Math Word Problems. Our model did not employ any special mechanisms or features specially designed for math word problems such as equation template extraction and multi-type attention. This highlights the importance of having our Graph2Tree neural networks that can leverage the structured information from both inputs and outputs for automatic solving of math problems.

%\footnotetext[1]{These methods are compared in \cite{robaidek2018data}.}
%\footnotetext[2]{We run our own implementation of Graph2Seq on thesedatasets using PyTorch.}

% \begin{table}[!ht]
% \scriptsize
% \centering
% \vspace{1mm}
% % \caption{Different equation generation strategies.}
% \newcommand{\Bd}[1]{\textbf{#1}}
% \vspace{-2mm}
% % \setlength{\tabcolsep}{7mm}{
% \begin{tabular}{|p{2.8in}|}
% \hline
% \textbf{Problem:} 0.5 of the cows are grazing grass .  0.25 of the cows are sleeping and 9 cows are drinking water from the pond .  find the total number of cows .\\
% \hline
% \textbf{w/o EN:} ( ( 0.5 * x ) + ( 0.25 * x ) ) + 9.0 = x\\
% \hline
% \textbf{w/ EN:} x = 9.0 / ( 1 - 0.5 - 0.25)\\
% \hline
% \end{tabular}
% \vspace{-2mm}
% % }
% \caption{\label{tab:expression_cmp} Different equation generation strategies.}
% \end{table}

% \noindent\textbf{Equation generation strategy.}
It is worth noting that our hierarchical tree decoder directly generates original mathematical expressions, which faithfully reflect reasoning steps when building math equations. However, state-of-the-art math word problem solvers like Group-Att \cite{li-etal-2019-modeling} or T-RNN \cite{Wang2019TemplateBasedMW} have achieved high performance by utilizing Equation Normalization (EN) proposed by \cite{Wang2019TemplateBasedMW} to keep structures of output equations unified. This method can improve solution accuracy because it reduces the difficulty of equation generation. On the other hand, the normalized equations completely lose the semantic meaning of operands and operators, making them difficult to reason rationales how answer math equations are built. 

\noindent\textbf{Attention visualization.}
For better understanding of our separated attention, we give a visualization sample from MAWPS.
As shown in Figure \ref{fig:att1}, we give an augmented graph input and equation tree, where $ \langle \textit{N}\rangle $ is sub-tree node and $ \textit{1},\textit{2} $ are indexed markers for original numbers. Specifically, Figure \ref{fig:att2} and \ref{fig:att3} illustrates alignments with word nodes and compositional nodes in graph input respectively. For example, in Figure \ref{fig:att3}, the equation part ``2 * 1'' is matched with ``a bee has 2 legs'' in the original natural language sentence which is actually semantically connected with ``NP'' and ``VP'' in the constituency tree. 

\begin{table}[ht!]
\centering
\vspace{-0mm}
\newcommand{\Bd}[1]{\textbf{#1}}
\begin{tabular}{|l||c|}
\hline
\Bd{Methods} & \Bd{MAWAPS} \\
\hline
Full model & 78.8\\\hline
w/o const tree & 75.6 \\
\hline
w/ original GraphSage & 76.4 \\ \hline
w/ only parent feeding & 75.6 \\ \hline
w/ only sibling feeding & 72.4 \\ \hline
w/o parent \& sibling feeding & 67.6 \\ \hline

w/o separated attention & 67.6 \\
\multicolumn{1}{|c||}{w/ uniform attention} & 71.6 \\
\hline
w/o bilstm & 72.8\\
\hline
\end{tabular}
\vspace{-0mm}
\caption{\label{tab:ablation study for mawps} Ablation study of Graph2Tree on the math word problem (MAWAPS). We employ solution accuracy as evaluation metric. The \textit{\textbf{Methods}} settings is same as Table  \ref{tab:ablation study}.} 
\end{table}

\noindent\textbf{Ablation study.}
Similarly, we also perform the ablation study for math word problem (MAWAPS), as shown in Table \ref{tab:ablation study for mawps}. This is done with dev set. Attention mechanism, constituency structure, and other components in our model play significant roles for Graph2tree to achieve high performance in MWP solving, which is consistent with our observation in the semantic parsing task. However, it is worth noting that, according to the experiment, the sibling mechanism is obviously more important to the MWP task than the semantic parsing task, which is in line with our expectations. In the MWP task, the result of decoding, math expressions, is relatively simple compared to semantic parsing. And in math expressions, the order between leaf nodes (numbers), which directly affects the correctness of expressions, is very important. The sibling mechanism plays exactly such a role. One potential interesting extension is that, if we can connect leaf nodes in the input graph and employ edge weights to dynamically represent the order between the nodes, it may achieve a similar or even better effect than the sibling mechanism. %This idea is reserved for future work.

\section{Conclusion and Future Work}
\label{sec:con}
% We presented a novel Graph2Tree model consisting of a graph encoder and a hierarchical tree decoder, for learning the translation between structured inputs and structured outputs. We studied the effectiveness of our Graph2Tree model on two tasks - Semantic Parsing and Math Word Problem and demonstrated that our model consistently outperformed or matched the performance of state-of-the-art baselines. We plan to develop a general Graph2Graph model as next future work.
We presented a novel Graph2Tree model consisting of a graph encoder and a hierarchical tree decoder, for learning the translation between structured inputs and structured outputs. Studies on two tasks - Semantic Parsing and Math Word Problem demonstrated our model consistently outperformed or matched the performance of the state-of-the-art. Our Graph2Tree model is generic and agnostic to the downstream tasks and thus one of the future works is to adapt it to the other NLP applications.

\bibliography{anthology,emnlp2020}

\begin{thebibliography}{56}
\expandafter\ifx\csname natexlab\endcsname\relax\def\natexlab#1{#1}\fi

\bibitem[{Altun et~al.(2004)Altun, Hofmann, and Smola}]{altun2004gaussian}
Yasemin Altun, Thomas Hofmann, and Alexander~J Smola. 2004.
\newblock Gaussian process classification for segmenting and annotating
  sequences.
\newblock In \emph{ICML}, page~4. ACM.

\bibitem[{Alvarez-Melis and Jaakkola(2017)}]{alvarez2016tree}
David Alvarez-Melis and Tommi~S Jaakkola. 2017.
\newblock Tree-structured decoding with doubly-recurrent neural networks.
\newblock \emph{ICLR}.

\bibitem[{Amini et~al.(2019)Amini, Gabriel, Lin, Koncel-Kedziorski, Choi, and
  Hajishirzi}]{amini-etal-2019-mathqa}
Aida Amini, Saadia Gabriel, Shanchuan Lin, Rik Koncel-Kedziorski, Yejin Choi,
  and Hannaneh Hajishirzi. 2019.
\newblock \href {https://doi.org/10.18653/v1/N19-1245} {{M}ath{QA}: Towards
  interpretable math word problem solving with operation-based formalisms}.
\newblock In \emph{Proceedings of the 2019 Conference of the North {A}merican
  Chapter of the Association for Computational Linguistics: Human Language
  Technologies, Volume 1 (Long and Short Papers)}, pages 2357--2367,
  Minneapolis, Minnesota. Association for Computational Linguistics.

\bibitem[{Bahdanau et~al.(2014)Bahdanau, Cho, and
  Bengio}]{bahdanau+al-2014-nmt}
Dzmitry Bahdanau, Kyunghyun Cho, and Yoshua Bengio. 2014.
\newblock \href {https://arxiv.org/abs/1409.0473} {Neural machine translation
  by jointly learning to align and translate}.
\newblock \emph{arXiv e-prints}, abs/1409.0473.

\bibitem[{Bastings et~al.(2017)Bastings, Titov, Aziz, Marcheggiani, and
  Sima{'}an}]{bastings-etal-2017-graph}
Joost Bastings, Ivan Titov, Wilker Aziz, Diego Marcheggiani, and Khalil
  Sima{'}an. 2017.
\newblock \href {https://doi.org/10.18653/v1/D17-1209} {Graph convolutional
  encoders for syntax-aware neural machine translation}.
\newblock In \emph{Proceedings of the 2017 Conference on Empirical Methods in
  Natural Language Processing}, pages 1957--1967, Copenhagen, Denmark.
  Association for Computational Linguistics.

\bibitem[{Beck et~al.(2018)Beck, Haffari, and Cohn}]{beck-etal-2018-graph}
Daniel Beck, Gholamreza Haffari, and Trevor Cohn. 2018.
\newblock \href {https://doi.org/10.18653/v1/P18-1026} {Graph-to-sequence
  learning using gated graph neural networks}.
\newblock In \emph{Proceedings of the 56th Annual Meeting of the Association
  for Computational Linguistics (Volume 1: Long Papers)}, pages 273--283,
  Melbourne, Australia. Association for Computational Linguistics.

\bibitem[{Chen et~al.(2018{\natexlab{a}})Chen, Sun, and
  Han}]{chen-etal-2018-sequence}
Bo~Chen, Le~Sun, and Xianpei Han. 2018{\natexlab{a}}.
\newblock \href {https://doi.org/10.18653/v1/P18-1071} {Sequence-to-action:
  End-to-end semantic graph generation for semantic parsing}.
\newblock In \emph{Proceedings of the 56th Annual Meeting of the Association
  for Computational Linguistics (Volume 1: Long Papers)}, pages 766--777,
  Melbourne, Australia. Association for Computational Linguistics.

\bibitem[{Chen et~al.(2019{\natexlab{a}})Chen, Huang, Palangi, Smolensky,
  Forbus, and Gao}]{chen2019natural}
Kezhen Chen, Qiuyuan Huang, Hamid Palangi, Paul Smolensky, Kenneth~D Forbus,
  and Jianfeng Gao. 2019{\natexlab{a}}.
\newblock Natural-to formal-language generation using tensor product
  representations.
\newblock \emph{arXiv preprint arXiv:1910.02339}.

\bibitem[{Chen et~al.(2018{\natexlab{b}})Chen, Liu, and Song}]{chen2018tree}
Xinyun Chen, Chang Liu, and Dawn Song. 2018{\natexlab{b}}.
\newblock Tree-to-tree neural networks for program translation.
\newblock In \emph{NIPS}, pages 2547--2557.

\bibitem[{Chen et~al.(2019{\natexlab{b}})Chen, Wu, and
  Zaki}]{chen2019graphflow}
Yu~Chen, Lingfei Wu, and Mohammed~J Zaki. 2019{\natexlab{b}}.
\newblock Graphflow: Exploiting conversation flow with graph neural networks
  for conversational machine comprehension.
\newblock \emph{arXiv preprint arXiv:1908.00059}.

\bibitem[{Chen et~al.(2020)Chen, Wu, and Zaki}]{chen2019reinforcement}
Yu~Chen, Lingfei Wu, and Mohammed~J Zaki. 2020.
\newblock Reinforcement learning based graph-to-sequence model for natural
  question generation.
\newblock \emph{ICLR}.

\bibitem[{Dahl et~al.(1994)Dahl, Bates, Brown, Fisher, Hunicke-Smith, Pallett,
  Pao, Rudnicky, and Shriberg}]{dahl-etal-1994-expanding}
Deborah~A. Dahl, Madeleine Bates, Michael Brown, William Fisher, Kate
  Hunicke-Smith, David Pallett, Christine Pao, Alexander Rudnicky, and
  Elizabeth Shriberg. 1994.
\newblock \href {https://www.aclweb.org/anthology/H94-1010} {Expanding the
  scope of the atis task: The atis-3 corpus}.
\newblock In \emph{{HUMAN} {LANGUAGE} {TECHNOLOGY}: Proceedings of a Workshop
  held at Plainsboro, New Jersey, March 8-11, 1994}.

\bibitem[{Dong and Lapata(2016)}]{dong-lapata-2016-language}
Li~Dong and Mirella Lapata. 2016.
\newblock \href {https://doi.org/10.18653/v1/P16-1004} {Language to logical
  form with neural attention}.
\newblock In \emph{Proceedings of the 54th Annual Meeting of the Association
  for Computational Linguistics (Volume 1: Long Papers)}, pages 33--43, Berlin,
  Germany. Association for Computational Linguistics.

\bibitem[{Dong and Lapata(2018)}]{dong-lapata-2018-coarse}
Li~Dong and Mirella Lapata. 2018.
\newblock \href {https://doi.org/10.18653/v1/P18-1068} {Coarse-to-fine decoding
  for neural semantic parsing}.
\newblock In \emph{Proceedings of the 56th Annual Meeting of the Association
  for Computational Linguistics (Volume 1: Long Papers)}, pages 731--742,
  Melbourne, Australia. Association for Computational Linguistics.

\bibitem[{Gilmer et~al.(2017)Gilmer, Schoenholz, Riley, Vinyals, and
  Dahl}]{gilmer2017neural}
Justin Gilmer, Samuel~S Schoenholz, Patrick~F Riley, Oriol Vinyals, and
  George~E Dahl. 2017.
\newblock Neural message passing for quantum chemistry.
\newblock In \emph{Proceedings of the 34th International Conference on Machine
  Learning-Volume 70}, pages 1263--1272. JMLR. org.

\bibitem[{Graves et~al.(2013)Graves, Mohamed, and Hinton}]{graves2013speech}
Alex Graves, Abdel-rahman Mohamed, and Geoffrey Hinton. 2013.
\newblock Speech recognition with deep recurrent neural networks.
\newblock In \emph{2013 IEEE international conference on acoustics, speech and
  signal processing}, pages 6645--6649. IEEE.

\bibitem[{G{\=u} et~al.(2018)G{\=u}, Shavarani, and Sarkar}]{gu-etal-2018-top}
Jetic G{\=u}, Hassan~S. Shavarani, and Anoop Sarkar. 2018.
\newblock \href {https://doi.org/10.18653/v1/D18-1037} {Top-down tree
  structured decoding with syntactic connections for neural machine translation
  and parsing}.
\newblock In \emph{Proceedings of the 2018 Conference on Empirical Methods in
  Natural Language Processing}, pages 401--413, Brussels, Belgium. Association
  for Computational Linguistics.

\bibitem[{Guo et~al.(2018)Guo, Wu, and Zhao}]{guo2018deep}
Xiaojie Guo, Lingfei Wu, and Liang Zhao. 2018.
\newblock Deep graph translation.
\newblock \emph{arXiv preprint arXiv:1805.09980}.

\bibitem[{Hamilton et~al.(2017)Hamilton, Ying, and
  Leskovec}]{hamilton2017inductive}
Will Hamilton, Zhitao Ying, and Jure Leskovec. 2017.
\newblock Inductive representation learning on large graphs.
\newblock In \emph{Advances in Neural Information Processing Systems}, pages
  1024--1034.

\bibitem[{Hosseini et~al.(2014)Hosseini, Hajishirzi, Etzioni, and
  Kushman}]{hosseini-etal-2014-learning}
Mohammad~Javad Hosseini, Hannaneh Hajishirzi, Oren Etzioni, and Nate Kushman.
  2014.
\newblock \href {https://doi.org/10.3115/v1/D14-1058} {Learning to solve
  arithmetic word problems with verb categorization}.
\newblock In \emph{Proceedings of the 2014 Conference on Empirical Methods in
  Natural Language Processing ({EMNLP})}, pages 523--533, Doha, Qatar.
  Association for Computational Linguistics.

\bibitem[{Jia and Liang(2016)}]{jia-liang-2016-data}
Robin Jia and Percy Liang. 2016.
\newblock \href {https://doi.org/10.18653/v1/P16-1002} {Data recombination for
  neural semantic parsing}.
\newblock In \emph{Proceedings of the 54th Annual Meeting of the Association
  for Computational Linguistics (Volume 1: Long Papers)}, pages 12--22, Berlin,
  Germany. Association for Computational Linguistics.

\bibitem[{Jie and Lu(2018)}]{jie-lu-2018-dependency}
Zhanming Jie and Wei Lu. 2018.
\newblock \href {https://doi.org/10.18653/v1/D18-1265} {Dependency-based hybrid
  trees for semantic parsing}.
\newblock In \emph{Proceedings of the 2018 Conference on Empirical Methods in
  Natural Language Processing}, pages 2431--2441, Brussels, Belgium.
  Association for Computational Linguistics.

\bibitem[{Joachims et~al.(2009)Joachims, Hofmann, Yue, and
  Yu}]{joachims2009predicting}
Thorsten Joachims, Thomas Hofmann, Yisong Yue, and Chun-Nam Yu. 2009.
\newblock Predicting structured objects with support vector machines.
\newblock \emph{Communications of the ACM}, 52(11):97.

\bibitem[{Kingma and Ba(2014)}]{kingma2014adam}
Diederik~P Kingma and Jimmy Ba. 2014.
\newblock Adam: A method for stochastic optimization.
\newblock \emph{arXiv preprint arXiv:1412.6980}.

\bibitem[{Kipf and Welling(2017)}]{kipf2017semi}
Thomas~N. Kipf and Max Welling. 2017.
\newblock Semi-supervised classification with graph convolutional networks.
\newblock In \emph{International Conference on Learning Representations
  (ICLR)}.

\bibitem[{Koncel-Kedziorski et~al.(2016)Koncel-Kedziorski, Roy, Amini, Kushman,
  and Hajishirzi}]{koncel-kedziorski-etal-2016-mawps}
Rik Koncel-Kedziorski, Subhro Roy, Aida Amini, Nate Kushman, and Hannaneh
  Hajishirzi. 2016.
\newblock \href {https://doi.org/10.18653/v1/N16-1136} {{MAWPS}: A math word
  problem repository}.
\newblock In \emph{Proceedings of the 2016 Conference of the North {A}merican
  Chapter of the Association for Computational Linguistics: Human Language
  Technologies}, pages 1152--1157, San Diego, California. Association for
  Computational Linguistics.

\bibitem[{Kushman et~al.(2014)Kushman, Artzi, Zettlemoyer, and
  Barzilay}]{kushman-etal-2014-learning}
Nate Kushman, Yoav Artzi, Luke Zettlemoyer, and Regina Barzilay. 2014.
\newblock \href {https://doi.org/10.3115/v1/P14-1026} {Learning to
  automatically solve algebra word problems}.
\newblock In \emph{Proceedings of the 52nd Annual Meeting of the Association
  for Computational Linguistics (Volume 1: Long Papers)}, pages 271--281,
  Baltimore, Maryland. Association for Computational Linguistics.

\bibitem[{Li et~al.(2018)Li, Tu, Yang, Lyu, and
  Zhang}]{li-etal-2018-multi-head}
Jian Li, Zhaopeng Tu, Baosong Yang, Michael~R. Lyu, and Tong Zhang. 2018.
\newblock \href {https://doi.org/10.18653/v1/D18-1317} {Multi-head attention
  with disagreement regularization}.
\newblock In \emph{Proceedings of the 2018 Conference on Empirical Methods in
  Natural Language Processing}, pages 2897--2903, Brussels, Belgium.
  Association for Computational Linguistics.

\bibitem[{Li et~al.(2019)Li, Wang, Zhang, Wang, Dai, and
  Zhang}]{li-etal-2019-modeling}
Jierui Li, Lei Wang, Jipeng Zhang, Yan Wang, Bing~Tian Dai, and Dongxiang
  Zhang. 2019.
\newblock \href {https://doi.org/10.18653/v1/P19-1619} {Modeling intra-relation
  in math word problems with different functional multi-head attentions}.
\newblock In \emph{Proceedings of the 57th Annual Meeting of the Association
  for Computational Linguistics}, pages 6162--6167, Florence, Italy.
  Association for Computational Linguistics.

\bibitem[{Li et~al.(2016)Li, Zemel, Brockschmidt, and Tarlow}]{li2016gated}
Yujia Li, Richard Zemel, Marc Brockschmidt, and Daniel Tarlow. 2016.
\newblock \href
  {https://www.microsoft.com/en-us/research/publication/gated-graph-sequence-neural-networks/}
  {Gated graph sequence neural networks}.
\newblock In \emph{Proceedings of ICLR'16}.

\bibitem[{Ling et~al.(2016)Ling, Blunsom, Grefenstette, Hermann,
  Ko{\v{c}}isk{\'y}, Wang, and Senior}]{ling-etal-2016-latent}
Wang Ling, Phil Blunsom, Edward Grefenstette, Karl~Moritz Hermann,
  Tom{\'a}{\v{s}} Ko{\v{c}}isk{\'y}, Fumin Wang, and Andrew Senior. 2016.
\newblock \href {https://doi.org/10.18653/v1/P16-1057} {Latent predictor
  networks for code generation}.
\newblock In \emph{Proceedings of the 54th Annual Meeting of the Association
  for Computational Linguistics (Volume 1: Long Papers)}, pages 599--609,
  Berlin, Germany. Association for Computational Linguistics.

\bibitem[{Ling et~al.(2017)Ling, Yogatama, Dyer, and
  Blunsom}]{ling-etal-2017-program}
Wang Ling, Dani Yogatama, Chris Dyer, and Phil Blunsom. 2017.
\newblock \href {https://doi.org/10.18653/v1/P17-1015} {Program induction by
  rationale generation: Learning to solve and explain algebraic word problems}.
\newblock In \emph{Proceedings of the 55th Annual Meeting of the Association
  for Computational Linguistics (Volume 1: Long Papers)}, pages 158--167,
  Vancouver, Canada. Association for Computational Linguistics.

\bibitem[{Luong et~al.(2015)Luong, Pham, and
  Manning}]{luong-etal-2015-effective}
Thang Luong, Hieu Pham, and Christopher~D. Manning. 2015.
\newblock \href {https://doi.org/10.18653/v1/D15-1166} {Effective approaches to
  attention-based neural machine translation}.
\newblock In \emph{Proceedings of the 2015 Conference on Empirical Methods in
  Natural Language Processing}, pages 1412--1421, Lisbon, Portugal. Association
  for Computational Linguistics.

\bibitem[{Manning et~al.(2014)Manning, Surdeanu, Bauer, Finkel, Bethard, and
  McClosky}]{manning-etal-2014-stanford}
Christopher Manning, Mihai Surdeanu, John Bauer, Jenny Finkel, Steven Bethard,
  and David McClosky. 2014.
\newblock \href {https://doi.org/10.3115/v1/P14-5010} {The {S}tanford
  {C}ore{NLP} natural language processing toolkit}.
\newblock In \emph{Proceedings of 52nd Annual Meeting of the Association for
  Computational Linguistics: System Demonstrations}, pages 55--60, Baltimore,
  Maryland. Association for Computational Linguistics.

\bibitem[{Peng et~al.(2018)Peng, Gildea, and Satta}]{peng2018amr}
Xiaochang Peng, Daniel Gildea, and Giorgio Satta. 2018.
\newblock Amr parsing with cache transition systems.
\newblock In \emph{Thirty-Second AAAI Conference on Artificial Intelligence}.

\bibitem[{Pennington et~al.(2014)Pennington, Socher, and
  Manning}]{pennington-etal-2014-glove}
Jeffrey Pennington, Richard Socher, and Christopher Manning. 2014.
\newblock \href {https://doi.org/10.3115/v1/D14-1162} {{G}love: Global vectors
  for word representation}.
\newblock In \emph{Proceedings of the 2014 Conference on Empirical Methods in
  Natural Language Processing ({EMNLP})}, pages 1532--1543, Doha, Qatar.
  Association for Computational Linguistics.

\bibitem[{Rabinovich et~al.(2017)Rabinovich, Stern, and
  Klein}]{rabinovich-etal-2017-abstract}
Maxim Rabinovich, Mitchell Stern, and Dan Klein. 2017.
\newblock \href {https://doi.org/10.18653/v1/P17-1105} {Abstract syntax
  networks for code generation and semantic parsing}.
\newblock In \emph{Proceedings of the 55th Annual Meeting of the Association
  for Computational Linguistics (Volume 1: Long Papers)}, pages 1139--1149,
  Vancouver, Canada. Association for Computational Linguistics.

\bibitem[{Reddy et~al.(2016)Reddy, T{\"a}ckstr{\"o}m, Collins, Kwiatkowski,
  Das, Steedman, and Lapata}]{reddy-etal-2016-transforming}
Siva Reddy, Oscar T{\"a}ckstr{\"o}m, Michael Collins, Tom Kwiatkowski, Dipanjan
  Das, Mark Steedman, and Mirella Lapata. 2016.
\newblock \href {https://doi.org/10.1162/tacl_a_00088} {Transforming dependency
  structures to logical forms for semantic parsing}.
\newblock \emph{Transactions of the Association for Computational Linguistics},
  4:127--140.

\bibitem[{Robaidek et~al.(2018)Robaidek, Koncel-Kedziorski, and
  Hajishirzi}]{robaidek2018data}
Benjamin Robaidek, Rik Koncel-Kedziorski, and Hannaneh Hajishirzi. 2018.
\newblock Data-driven methods for solving algebra word problems.
\newblock \emph{arXiv preprint arXiv:1804.10718}.

\bibitem[{Song et~al.(2018)Song, Zhang, Wang, and
  Gildea}]{song-etal-2018-graph}
Linfeng Song, Yue Zhang, Zhiguo Wang, and Daniel Gildea. 2018.
\newblock \href {https://doi.org/10.18653/v1/P18-1150} {A graph-to-sequence
  model for {AMR}-to-text generation}.
\newblock In \emph{Proceedings of the 56th Annual Meeting of the Association
  for Computational Linguistics (Volume 1: Long Papers)}, pages 1616--1626,
  Melbourne, Australia. Association for Computational Linguistics.

\bibitem[{Sutskever et~al.(2014)Sutskever, Vinyals, and
  Le}]{sutskever2014sequence}
Ilya Sutskever, Oriol Vinyals, and Quoc~V Le. 2014.
\newblock Sequence to sequence learning with neural networks.
\newblock In \emph{NIPS}, pages 3104--3112.

\bibitem[{Tsochantaridis et~al.(2005)Tsochantaridis, Joachims, Hofmann, and
  Altun}]{tsochantaridis2005large}
Ioannis Tsochantaridis, Thorsten Joachims, Thomas Hofmann, and Yasemin Altun.
  2005.
\newblock Large margin methods for structured and interdependent output
  variables.
\newblock \emph{Journal of machine learning research}, 6(Sep):1453--1484.

\bibitem[{Velickovic et~al.(2017)Velickovic, Cucurull, Casanova, Romero,
  Li{\`o}, and Bengio}]{Velickovic2017GraphAN}
Petar Velickovic, Guillem Cucurull, Arantxa Casanova, Adriana Romero, Pietro
  Li{\`o}, and Yoshua Bengio. 2017.
\newblock Graph attention networks.
\newblock \emph{ArXiv}, abs/1710.10903.

\bibitem[{Wang et~al.(2018{\natexlab{a}})Wang, Wang, Cai, Zhang, and
  Liu}]{wang-etal-2018-translating}
Lei Wang, Yan Wang, Deng Cai, Dongxiang Zhang, and Xiaojiang Liu.
  2018{\natexlab{a}}.
\newblock \href {https://doi.org/10.18653/v1/D18-1132} {Translating a math word
  problem to a expression tree}.
\newblock In \emph{Proceedings of the 2018 Conference on Empirical Methods in
  Natural Language Processing}, pages 1064--1069, Brussels, Belgium.
  Association for Computational Linguistics.

\bibitem[{Wang et~al.(2018{\natexlab{b}})Wang, Zhang, Gao, Song, Guo, and
  Shen}]{wang2018mathdqn}
Lei Wang, Dongxiang Zhang, Lianli Gao, Jingkuan Song, Long Guo, and Heng~Tao
  Shen. 2018{\natexlab{b}}.
\newblock Mathdqn: Solving arithmetic word problems via deep reinforcement
  learning.
\newblock In \emph{AAAI}.

\bibitem[{Wang et~al.(2019)Wang, Zhang, Zhang, Xu, Gao, Dai, and
  Shen}]{Wang2019TemplateBasedMW}
Lei Wang, Dongxiang Zhang, Jipeng Zhang, Xing Xu, Lianli Gao, Bing~Tian Dai,
  and Heng~Tao Shen. 2019.
\newblock Template-based math word problem solvers with recursive neural
  networks.
\newblock In \emph{AAAI}.

\bibitem[{Wang et~al.(2017)Wang, Lu, Zhou, and Liu}]{wang-etal-2017-deep}
Mingxuan Wang, Zhengdong Lu, Jie Zhou, and Qun Liu. 2017.
\newblock \href {https://doi.org/10.18653/v1/P17-1013} {Deep neural machine
  translation with linear associative unit}.
\newblock In \emph{Proceedings of the 55th Annual Meeting of the Association
  for Computational Linguistics (Volume 1: Long Papers)}, pages 136--145,
  Vancouver, Canada. Association for Computational Linguistics.

\bibitem[{Xie and Sun(2019)}]{ijcai2019-736}
Zhipeng Xie and Shichao Sun. 2019.
\newblock \href {https://doi.org/10.24963/ijcai.2019/736} {A goal-driven
  tree-structured neural model for math word problems}.
\newblock In \emph{Proceedings of the Twenty-Eighth International Joint
  Conference on Artificial Intelligence, {IJCAI-19}}, pages 5299--5305.
  International Joint Conferences on Artificial Intelligence Organization.

\bibitem[{Xu et~al.(2018{\natexlab{a}})Xu, Wu, Wang, and
  Sheinin}]{xu2018graph2seq}
Kun Xu, Lingfei Wu, Zhiguo Wang, and Vadim Sheinin. 2018{\natexlab{a}}.
\newblock Graph2seq: Graph to sequence learning with attention-based neural
  networks.
\newblock \emph{arXiv preprint arXiv:1804.00823}.

\bibitem[{Xu et~al.(2018{\natexlab{b}})Xu, Wu, Wang, Yu, Chen, and
  Sheinin}]{xu2018exploiting}
Kun Xu, Lingfei Wu, Zhiguo Wang, Mo~Yu, Liwei Chen, and Vadim Sheinin.
  2018{\natexlab{b}}.
\newblock Exploiting rich syntactic information for semantic parsing with
  graph-to-sequence model.
\newblock \emph{arXiv preprint arXiv:1808.07624}.

\bibitem[{Xu et~al.(2018{\natexlab{c}})Xu, Wu, Wang, Yu, Chen, and
  Sheinin}]{xu2018sql}
Kun Xu, Lingfei Wu, Zhiguo Wang, Mo~Yu, Liwei Chen, and Vadim Sheinin.
  2018{\natexlab{c}}.
\newblock Sql-to-text generation with graph-to-sequence model.
\newblock \emph{arXiv preprint arXiv:1809.05255}.

\bibitem[{Yin and Neubig(2017)}]{yin-neubig-2017-syntactic}
Pengcheng Yin and Graham Neubig. 2017.
\newblock \href {https://doi.org/10.18653/v1/P17-1041} {A syntactic neural
  model for general-purpose code generation}.
\newblock In \emph{Proceedings of the 55th Annual Meeting of the Association
  for Computational Linguistics (Volume 1: Long Papers)}, pages 440--450,
  Vancouver, Canada. Association for Computational Linguistics.

\bibitem[{Yin et~al.(2019)Yin, Neubig, Allamanis, Brockschmidt, and
  Gaunt}]{YinNABG19}
Pengcheng Yin, Graham Neubig, Miltiadis Allamanis, Marc Brockschmidt, and
  Alexander~L. Gaunt. 2019.
\newblock \href {https://openreview.net/forum?id=BJl6AjC5F7} {Learning to
  represent edits}.
\newblock In \emph{7th International Conference on Learning Representations,
  {ICLR} 2019, New Orleans, LA, USA, May 6-9, 2019}.

\bibitem[{Yin et~al.(2018)Yin, Zhou, He, and Neubig}]{yin-etal-2018-structvae}
Pengcheng Yin, Chunting Zhou, Junxian He, and Graham Neubig. 2018.
\newblock \href {https://doi.org/10.18653/v1/P18-1070} {{S}truct{VAE}:
  Tree-structured latent variable models for semi-supervised semantic parsing}.
\newblock In \emph{Proceedings of the 56th Annual Meeting of the Association
  for Computational Linguistics (Volume 1: Long Papers)}, pages 754--765,
  Melbourne, Australia. Association for Computational Linguistics.

\bibitem[{Zettlemoyer and Collins(2005)}]{Zettlemoyer05learningto}
Luke~S. Zettlemoyer and Michael Collins. 2005.
\newblock Learning to map sentences to logical form: Structured classification
  with probabilistic categorial grammars.
\newblock In \emph{In Proceedings of the 21st Conference on Uncertainty in AI},
  pages 658--666.

\bibitem[{Zou and Lu(2019)}]{zou-lu-2019-text2math}
Yanyan Zou and Wei Lu. 2019.
\newblock \href {https://doi.org/10.18653/v1/D19-1536} {{T}ext2{M}ath:
  End-to-end parsing text into math expressions}.
\newblock In \emph{Proceedings of the 2019 Conference on Empirical Methods in
  Natural Language Processing and the 9th International Joint Conference on
  Natural Language Processing (EMNLP-IJCNLP)}, pages 5327--5337, Hong Kong,
  China. Association for Computational Linguistics.

\end{thebibliography}
\bibliographystyle{acl_natbib}

\end{document}

\maketitle

\appendix\section{Data Preprocessing and Computational Devices}
In both semantic parsing and math word problem experiments, we use CoreNLP Tool \cite{manning-etal-2014-stanford} as our external parser to generate constituency trees and dependency parsing trees. For structural brevity, we cut the root node in parsing trees and if nodes in graph are arranged into a "line" without branches, we transform them to a single node. In our experiments, we train our Graph2Tree model on a common single GPU: TITAN Xp. OS version: Ubuntu 16.04.4 LTS, and CUDA version: 8.0.

\section{Hyperparameter Setup}
The Graph2Tree model is trained with Adam optimizer \cite{kingma2014adam} with learning rate set to 0.001. We use ReLU as our non-linear function and the greedy search as our inference strategy for decoding. Word embedding layer is initialized with GloVe vectors(300 dimension) from \cite{pennington-etal-2014-glove}. The dropout rate is chosen from \{0.1, 0.3, 0.5\}.

For semantic parsing experiments, we have described parameter setting in Section \textbf{5.1}.

For math word problem experiments, we here give the details of experiments due to page limitation.

i) On \textbf{MAWPS} \cite{koncel-kedziorski-etal-2016-mawps} dataset, we randomly split the original dataset with the same train/dev/test split ratio following \cite{robaidek2018data}. To reduce vocabulary size, we projected numbers in problem descriptions to indexed markers, e.g., "n1, n2, \dots". The batch size we use is 30. The dropout rate is chosen from \{0.1,0.3\}. The hop size in GNN is chosen from \{2,3,4\}.

ii) On \textbf{MATHQA} \cite{amini-etal-2019-mathqa} dataset, the train/dev/test split is 29837/4475/2985. We use the annotated formula as our output. 
% Due to the lack of execution scripts, 
% we use exact match as criterion in our Graph2Tree and other baselines run by ourselves, which is a stricter criterion compared to solution accuracy. 
The batch size we use is 32. For other hyper-parameters, we use similar settings as in MAWPS.

\section{Sample Inputs and Outputs}
We show the generated results of Graph2Tree model in MAWPS dataset in \ref{table:smaples}. For generated math equations, we replace original numbers to indexed markers, e.g., \{1, 2, \dots\}.

\begin{table*}
\centering

\begin{tabular}{p{15.00cm}}
\hline
\textbf{Problem and math expression}\\
\hline\hline
what is 1 \% of 2 ? \\
\textit{Ground truth:} 1 * 0.01 * 2 = x\\
\textit{Graph2Tree:} x = ( 1 * 0.01 ) * 2 \\

\hline
twice a number increased by 2 is 1 . find the number .\\
\textit{Ground truth:} ( 2.0 * x ) + 2 = 1\\
\textit{Graph2Tree:} ( 2.0 * x ) + 2 = 1 \\

\hline
a number is 1 less than the sum of 2 and 3 . what is the number ?\\
\textit{Ground truth:} ( 2 + 3 ) - 1 = x\\
\textit{Graph2Tree:} ( x - 1 ) - 1 = ( 2 + 3 ) - 3 \\

\hline
4 times the sum of a number and 2 is 3 less than 1 times that number . what is the number ?\\
\textit{Ground truth:} 4 * ( 2 + x ) = ( 1 * x ) - 3\\
\textit{Graph2Tree:} 4 * ( 2 + x ) = ( 1 * x ) - 3 \\

\hline
in a class of 2 students , 1 received a grade of a . what percent of the students received a 's ?\\
\textit{Ground truth:} x = ( 1 / 2 ) * 100.0\\
\textit{Graph2Tree:} x = ( 1 / 2 ) * 100.0 \\

\hline
mrs . hilt bought a yoyo for 1 cents and a whistle for 2 cents . how much did she spend in all for the two toys ?\\
\textit{Ground truth:} x = ( 1 + 2 )\\
\textit{Graph2Tree:} 1 + 2 = x \\

\hline
isabella's hair is 2 inches long . if she gets a haircut and now her hair is 1 inches long , how much of isabella's hair got cut off ?\\
\textit{Ground truth:} x = ( 2 - 1 )\\
\textit{Graph2Tree:} x = ( 1 - 2 ) \\

\hline
if sally can paint a house in 1 hours , and john can paint the same house in 2 hour , how many hours will it take for both of them to paint the house together ?\\
\textit{Ground truth:} ( 1.0 / 1 ) + ( 1.0 / 2 ) = 1.0 / x\\
\textit{Graph2Tree:} ( ( 1.0 / 1 ) * x ) + ( ( 1.0 / 2 ) * x ) = 1.0 \\

\hline
alyssa went to 1 soccer games this year , but missed 2 . she went to 3 games last year and plans to go to 4 games next year . how many soccer games will alyssa go to in all ?\\
\textit{Ground truth:} x = 1 + 3 + 4\\
\textit{Graph2Tree:} x = 1 + 3 + 4 \\

\hline
a lawyer bills her clients 1 dollars per hour of service . if a client 's case requires 2 hours to complete , use proportion to calculate how much the client will owe the lawyer in dollars .\\
\textit{Ground truth:} 1 * 2 = x\\
\textit{Graph2Tree:} x = ( 2 / 1 )\\
\hline

\end{tabular}
\caption{Sample inputs and outputs of Graph2Tree on MAWPS.}
\label{table:smaples}
\end{table*}

\bibliography{anthology,emnlp2020}
\bibliographystyle{acl_natbib}